%% file: arxiv.tex
\documentclass[10pt,twocolumn,letterpaper]{article}

\usepackage[pagenumbers]{cvpr} 


\definecolor{cvprblue}{rgb}{0.21,0.49,0.74}
\usepackage[pagebackref,breaklinks,colorlinks,allcolors=cvprblue]{hyperref}

\usepackage{url}
\usepackage{graphicx}
\usepackage{colortbl}
\usepackage{amsmath}
\usepackage{amssymb}
\usepackage{amsfonts}
\usepackage{mathtools}
\usepackage{adjustbox}
\usepackage{marvosym}
\usepackage{multirow}
\usepackage{graphicx}
\usepackage{booktabs}
\usepackage{caption}
\usepackage{subcaption}
\usepackage{tabularx}
\usepackage{pifont}
\usepackage{arydshln}
\usepackage{makecell}
\usepackage{wrapfig}
\usepackage{float}
\usepackage{xspace}
\usepackage{array}

\usepackage[accsupp]{axessibility} 

\definecolor{backcolor}{RGB}{232, 242, 255}
\definecolor{green}{RGB}{0, 133, 21}
\definecolor{deemph}{gray}{0.6}

\newcommand{\names}{\textsc{EMOVA}\xspace}
\newcommand{\cmark}{\ding{51}}
\newcommand{\xmark}{\ding{55}}

\makeatletter
\def\blfootnote{\xdef\@thefnmark{}\@footnotetext}
\makeatother
 
\def\paperID{***} 
\def\confName{CVPR}
\def\confYear{2025}

\title{\includegraphics[width=1.25cm]{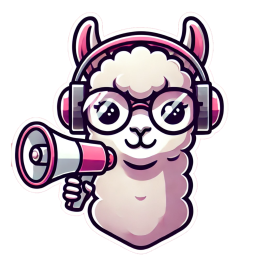} EMOVA: Empowering Language Models to See, Hear and Speak with \\ Vivid Emotions}

\author{Kai Chen$^{1*}$,
Yunhao Gou$^{1,6*}$,
Runhui Huang$^{2*}$,
Zhili Liu$^{1,3*}$,
Daxin Tan$^{3*}$,
Jing Xu$^{4}$,
Chunwei Wang$^{3}$,\\
Yi Zhu$^{3}$,
Yihan Zeng$^{3}$,
Kuo Yang$^{3}$,
Dingdong Wang$^{4}$,
Kun Xiang$^{5}$,
Haoyuan Li$^{5}$, 
Haoli Bai$^{3}$,\\
Jianhua Han$^{3}$,
Xiaohui Li$^{3}$,
Weike Jin$^{3}$,
Nian Xie$^{3}$, 
Yu Zhang$^{6}$,
James T. Kwok$^{1}$,\\
Hengshuang Zhao$^{2}$,
Xiaodan Liang$^{5}$,
Dit-Yan Yeung$^{1}$,
Xiao Chen$^{3}$,
Zhenguo Li$^{3}$,\\ 
Wei Zhang$^{3}$,
Qun Liu$^{3}$,
Jun Yao$^{3}$,
Lanqing Hong$^{3\dag}$,
Lu Hou$^{3\dag}$,
Hang Xu$^{3\dag}$
\vspace{0.125cm} \\
$^{1}$Hong Kong University of Science and Technology
\enspace
$^{2}$The University of Hong Kong\\
$^{3}$Huawei Noah's Ark Lab
\enspace
$^{4}$The Chinese University of Hong Kong\\
$^{5}$Sun Yat-sen University
\enspace
$^{6}$Southern University of Science and Technology
\vspace{0.125cm} \\
{\centering Project Page: \url{https://emova-ollm.github.io/}}
}

\begin{document}
\maketitle
\input{section_arxiv/0_abstract}
\input{section_arxiv/1_intro}
\input{section_arxiv/2_related_work}
\input{section_arxiv/3_arch}
\input{section_arxiv/4_alignment}
\input{section_arxiv/5_experiment}
\input{section_arxiv/6_conclusion}

\clearpage
\paragraph{Acknowledgments.}
We gratefully acknowledge supports of MindSpore, CANN (Compute Architecture for Neural Networks) and Ascend AI Processor used for this research.
This work has been made possible by a Research Impact Fund project (RIF R6003-21) and a General Research Fund project (GRF 16203224) funded by the Research Grants Council (RGC) of the Hong Kong Government.
This work was partially supported by the Research Grants Council of the Hong Kong Special Administrative Region (Grants C7004-22G-1 and 16202523) and the Joint Centre for Artificial Intelligence (Grant FB453).
This work is supported by National Key Research and Development Program of China (2024YFE0203100) and National Natural Science Foundation of China (No. 62441615, 62201484 and 62136005).


{
    \small
    \bibliographystyle{ieeenat_fullname}
    \bibliography{main}
}


\clearpage
\input{section_arxiv/7_appendix}

\end{document}

%% file: section_arxiv/0_abstract.tex
\begin{abstract}
GPT-4o, an omni-modal model that enables vocal conversations with diverse emotions and tones, marks a milestone for omni-modal foundation models.
However, empowering Large Language Models to perceive and generate images, texts, and speeches end-to-end with publicly available data remains challenging for the open-source community.
Existing vision-language models rely on external tools for speech processing, while speech-language models still suffer from limited or totally without vision-understanding capabilities.
To address this gap, we propose the \textbf{\names} (\textbf{\underline{EM}}otionally \textbf{\underline{O}}mni-present \textbf{\underline{V}}oice \textbf{\underline{A}}ssistant), to enable Large Language Models with end-to-end speech abilities while maintaining the leading vision-language performance.
With a semantic-acoustic disentangled speech tokenizer, we surprisingly notice that omni-modal alignment can further enhance vision-language and speech abilities compared with the bi-modal aligned counterparts.
Moreover, a lightweight style module is introduced for the flexible speech style controls including emotions and pitches.
For the first time, \textbf{\names} achieves state-of-the-art performance on both the vision-language and speech benchmarks, and meanwhile, supporting omni-modal spoken dialogue with vivid emotions. 
\end{abstract}

\blfootnote{$^{*}$Equal contribution, listed in alphabetical order by surname.}
\blfootnote{$^{\dag}$Corresponding authors: \texttt{\{honglanqing,houlu3,xu.hang\}@ huawei.com}}

%% file: section_arxiv/1_intro.tex
\vspace{-5mm}
\section{Introduction}
\label{sec:introduction}
\vspace{-1mm}

OpenAI GPT-4o~\citep{openai2024gpt4o}, a novel milestone for the omni-modal foundation models, has rekindled people's attention on intelligent assistants that can \textit{see} (\ie, perceiving fine-grained visual inputs), \textit{hear} (\ie, understanding vocal instructions) and \textit{speak} (\ie, generating vocal responses) simultaneously.
Existing Multi-modal Large Language Models (MLLMs) mostly focus on two modalities only, either vision-language \citep{bai2023qwen,li2024llavaonevisioneasyvisualtask} or speech-language~\citep{chu2024qwen2audiotechnicalreport,xie2024mini}, demonstrating severe demands for omni-modal models with visual, language and speech abilities.
\textit{How to empower Large Language Models (LLMs) to effectively process omni-modal data in an end-to-end manner remains an open question.}


\begin{figure}[t]
    \centering
    \vspace{-9mm}
    \includegraphics[width=0.95\linewidth]{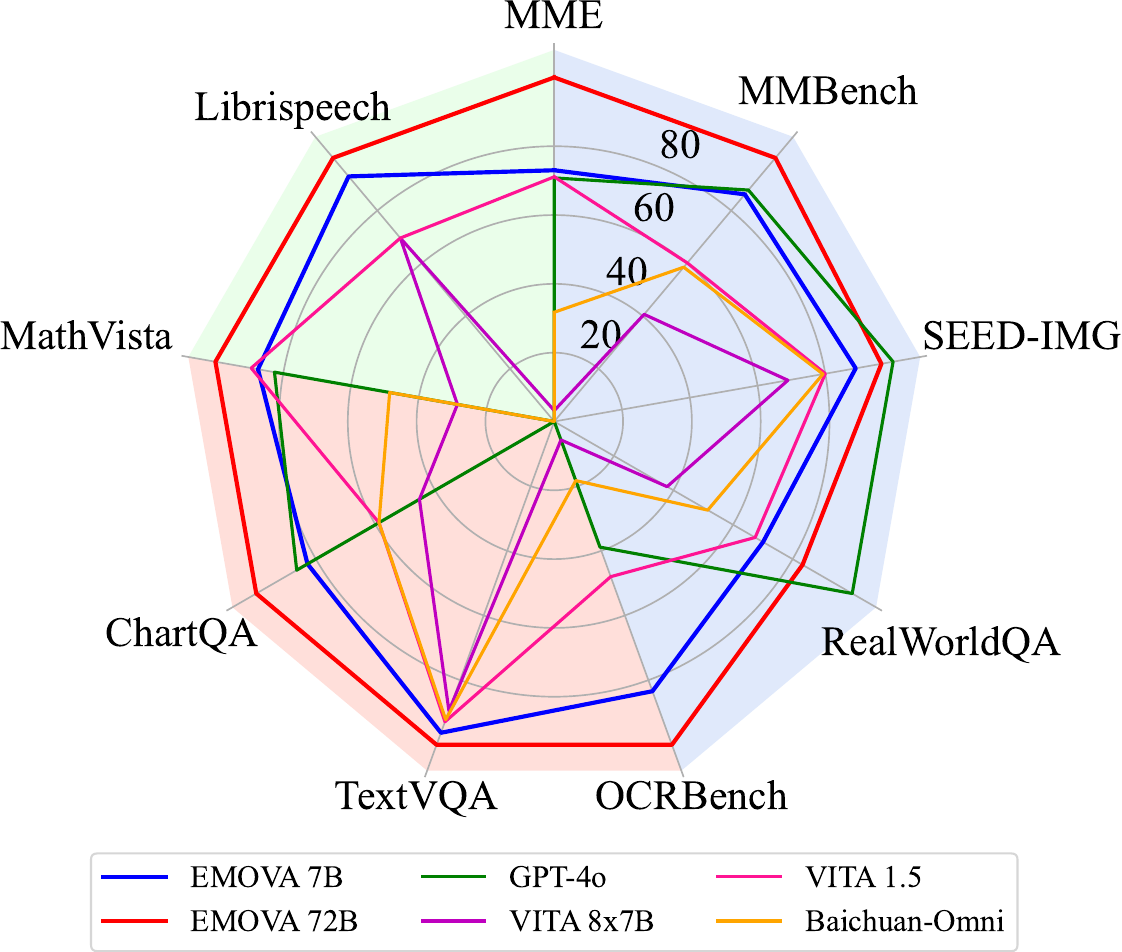}
    \vspace{-3mm}
    \caption{\textbf{\names} is the very first omni-modal LLM with state-of-the-art performance on both vision-language and speech benchmarks simultaneously. See detailed results in Table~\ref{tab:main_table}.}
    \label{fig:radar}
    \vspace{-6mm}
\end{figure}


Existing omni-modal LLMs~\citep{chen2024internvlscalingvisionfoundation,fu2024vitaopensourceinteractiveomni} generally build on 
Vision LLMs and integrate the speech modality by adopting a speech encoder like Whisper~\citep{radford2022robustspeechrecognitionlargescale}, which extracts \textit{continuous} speech features, similar to how images are processed, to enable speech understanding. 
These models, however, still rely on external Text-to-Speech (TTS) tools for generating speech responses, limiting their ability to support real-time interactions.
AnyGPT~\citep{zhan2024anygptunifiedmultimodalllm}, on the contrary, opts for a fully \textit{discretization} way, which first discretizes all data modalities (\ie, images, texts, and speeches), followed by omni-modal auto-regressive modeling. 
This enables AnyGPT to handle multiple modalities with a \textbf{unified end-to-end} framework, facilitating \textbf{real-time interactions} with the help of streaming decoding.
However, the discrete vision tokenizer used by AnyGPT struggles to capture visual details, especially for high-resolution images, falling far behind its continuous counterparts on vision-language benchmarks.
Furthermore, none of the existing works explore the speech style controls (\eg, emotions and pitches) with LLMs.
Thus, our question arises: \textit{How to build an end-to-end omni-modal LLM enabling emotional spoken dialogue while maintaining state-of-the-art vision-language performance?}

In this paper, we propose \textbf{\names} (\textbf{\underline{EM}}otionally \textbf{\underline{O}}mni-present \textbf{\underline{V}}oice \textbf{\underline{A}}ssistant), a novel end-to-end omni-modal LLM with state-of-the-art vision-language and speech capabilities while supporting emotional spoken dialogue.
Fig.~\ref{fig:framework} shows an overview of the model framework.
A \textit{continuous} vision encoder captures fine-grained visual details, while a \textit{discrete} speech tokenizer and detokenizer empower end-to-end speech understanding and generation. 
Specifically, the speech-to-unit (S2U) tokenizer tokenizes the input speech waveforms to discrete speech units, while the unit-to-speech (U2S) detokenizer reconstructs the speech waveforms from LLM's output speech units.
To seamlessly integrate speech modality with LLMs, we meticulously design a \textbf{semantic-} \textbf{acoustic disentangled} speech tokenizer to decouple the semantic contents and acoustic styles of input speeches \citep{tao2024toneunitspeechdiscretizationapproach}, where 1) \textit{semantic content} (\ie, what it says) captures the semantic meanings of the input speeches, which is finally discretized and aligned with LLMs, while
2) \textit{acoustic style} (\ie, how it says) captures the diverse speech styles (\eg, emotions and pitches).
Utilizing the semantic-acoustic disentanglement of our speech tokenizer, we further introduce a lightweight style module to support spoken dialogue with vivid emotions and pitches.
As in Sec.~\ref{sec_speech_in}, this disentanglement design better facilitates the modality alignment among texts and speeches while maintaining flexibility for diverse speech style controllability and personalization.

With \textbf{\names}'s end-to-end omni-modal framework, we empirically demonstrate publicly available bi-modal image-text and speech-text data are sufficient for the omni-modal alignment with the text modality as a bridge, eliminating the need for omni-modal data (\ie, image-text-speech), which is usually scarce. 
Surprisingly, we find that the omni-modal alignment can further improve both the vision-language and speech capabilities via joint optimization, even when compared with their bi-modal aligned counterparts.
Ultimately, only a small amount of omni-modality samples are required to teach the model to respond in the desired format.
For the first time, \textbf{\names} obtains state-of-the-art results on both vision-language and speech benchmarks (see Table~\ref{tab:main_table}).
The main contributions of this work contain three parts:
 
\begin{enumerate}
    \item We propose \textbf{\names}, a novel end-to-end omni-modal LLM that can see, hear and speak. 
    A continuous vision encoder with a semantic-acoustic disentangled speech tokenizer is adopted for seamless omni-modal alignment and diverse speech style controllability. 
    \item We introduce an efficient text-centric omni-modal alignment which
    can further enhance the vision-language and speech abilities, surpassing their bi-modal aligned counterparts 
    (\ie, image-text only and speech-text only).
    \item For the first time, \textbf{\names} obtains state-of-the-art comparable results on both the vision-language and speech benchmarks simultaneously and further supports flexible spoken dialogues with vivid emotions.
\end{enumerate}

%% file: section_arxiv/2_related_work.tex

\begin{table}[t]
\setlength{\tabcolsep}{1mm}
\begin{center}
\scalebox{0.9}{
\begin{tabular}{lccccc}
\toprule
\multirow{2}*{Method} & \multirow{2}*{Visual} & \multirow{2}*{\ \
Text\ \ } & \multicolumn{3}{c}{Speech} \\
& & & Understand & Gen. & Emotion \\
\midrule
\multicolumn{6}{l}{\textcolor{deemph}{\textit{Vision Large Language Models}}} \\
LLaVA~\cite{liu2023llava} & \cellcolor{green!10}\cmark & \cellcolor{green!10}\cmark & \cellcolor{red!10}\xmark & \cellcolor{red!10}\xmark & \cellcolor{red!10}\xmark \\
Intern-VL~\cite{chen2024internvlscalingvisionfoundation} & \cellcolor{green!10}\cmark & \cellcolor{green!10}\cmark & \cellcolor{red!10}\xmark & \cellcolor{red!10}\xmark & \cellcolor{red!10}\xmark \\
\midrule
\multicolumn{6}{l}{\textcolor{deemph}{\textit{Speech Large Language Models}}} \\
Qwen-Audio~\cite{chu2024qwen2audiotechnicalreport} & \cellcolor{red!10}\xmark & \cellcolor{green!10}\cmark & \cellcolor{green!10}\cmark & \cellcolor{red!10}\xmark & \cellcolor{red!10}\xmark \\
Mini-Omni~\cite{xie2024mini} & \cellcolor{red!10}\xmark & \cellcolor{green!10}\cmark & \cellcolor{green!10}\cmark & \cellcolor{green!10}\cmark & \cellcolor{red!10}\xmark \\
LLaMA-Omni~\cite{fang2024llamaomniseamlessspeechinteraction} & \cellcolor{red!10}\xmark & \cellcolor{green!10}\cmark & \cellcolor{green!10}\cmark & \cellcolor{green!10}\cmark & \cellcolor{red!10}\xmark \\
\midrule
\multicolumn{6}{l}{\textcolor{deemph}{\textit{Omni-modal Large Language Models}}} \\
VITA~\cite{fu2024vitaopensourceinteractiveomni,fu2025vita} & \cellcolor{green!10}\cmark & \cellcolor{green!10}\cmark & \cellcolor{green!10}\cmark & \cellcolor{red!10}\xmark & \cellcolor{red!10}\xmark \\
Ola~\cite{liu2025ola} & \cellcolor{green!10}\cmark & \cellcolor{green!10}\cmark & \cellcolor{green!10}\cmark & \cellcolor{red!10}\xmark & \cellcolor{red!10}\xmark \\
Any-GPT~\cite{zhan2024anygptunifiedmultimodalllm} & \cellcolor{green!10}\cmark & \cellcolor{green!10}\cmark & \cellcolor{green!10}\cmark & \cellcolor{green!10}\cmark & \cellcolor{red!10}\xmark \\
Baichuan-Omni~\cite{li2024baichuanomni} & \cellcolor{green!10}\cmark & \cellcolor{green!10}\cmark & \cellcolor{green!10}\cmark & \cellcolor{green!10}\cmark & \cellcolor{red!10}\xmark \\
\textbf{\names (ours)} & \cellcolor{green!10}\cmark & \cellcolor{green!10}\cmark & \cellcolor{green!10}\cmark & \cellcolor{green!10}\cmark & \cellcolor{green!10}\cmark \\
\bottomrule
\end{tabular}
}
\vspace{-3mm}
\caption{{\bf Comparison of Multi-modal Large Language Models.}
\textbf{\names} is the very first Omni-modal LLM capable of emotional spoken dialogue with state-of-the-art vision-language and speech capabilities simultaneously.
``Gen.'' stands for Generation.
}
\label{tab:compare}
\end{center}
\vspace{-9mm}
\end{table}


\vspace{-2mm}
\section{Related Work}
\label{sec:related work}
\vspace{-1mm}

\paragraph{Vision Large Language Models} (VLLMs) integrate the vision modality into LLMs~\citep{touvron2023llama,chen2023gaining}, enabling the advanced understanding and reasoning over visual instructions~\citep{liu2023llava,bai2023qwen,gou2023mixture,gou2024eyes}.
Recent VLLM works can be categorized into three directions, 1) \textit{Vision encoders}~\citep{oquab2023dinov2,chen2021multisiam,chen2023mixed} are enhanced and aggregated for robust representations~\citep{lin2023sphinx,li2024mini,tong2024cambrian}.
2) \textit{High-resolution} methods are proposed to overcome the fixed resolution of pre-trained vision encoders 
(e.g., $336 \times 336$ for CLIP~\citep{radford2021learningtransferablevisualmodels}), enabling LLMs to perceive fine-grained visual information~\citep{liu2024llavanext,dong2024xcomposer2-4khd,huang2024hires,luo2024feast}.
3) \textit{High-quality instruction data} is essential for VLLMs to generate accurate and well-formed responses~\citep{laurenccon2024matters,li2024llavaonevisioneasyvisualtask,chen2024internvlscalingvisionfoundation}.
Besides achieving state-of-the-art vision-language performance, we further introduce speech understanding and generating abilities to \textbf{\names}.


\begin{figure*}[t]
\centering
\includegraphics[width=0.8\textwidth]{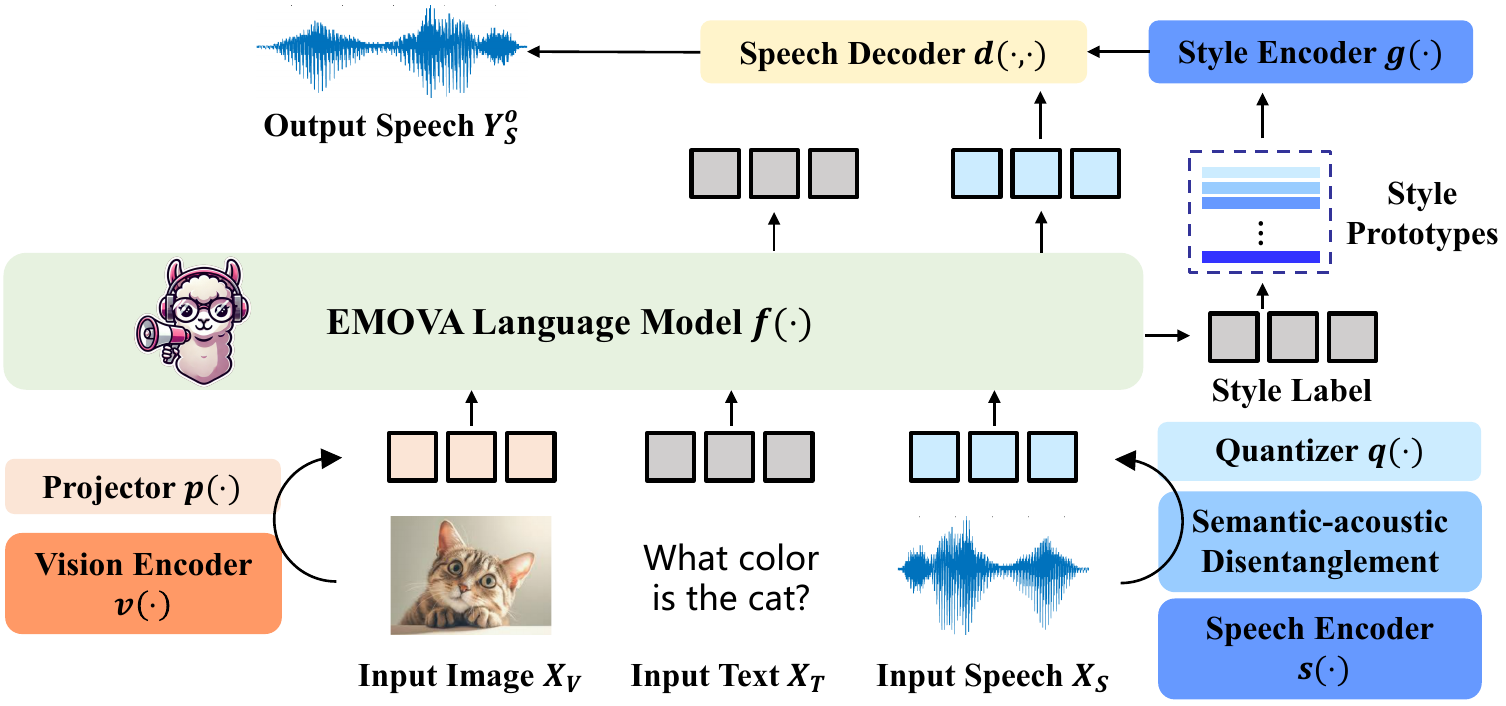}
\vspace{-3mm}
\caption{\textbf{Model architecture of \names.} 
The vision encoder extracts continuous visual features, which are projected into the textual embedding space as visual tokens, while the input speech is encoded and quantized into discrete speech units. 
Given the omni-modal inputs, \textbf{\names} can generate both textual and speech responses with vivid emotional controls. 
Check Sec.~\ref{sec:method} for more architectural details.
}
\vspace{-5mm}
\label{fig:framework}
\end{figure*}


\vspace{-2mm}
\paragraph{Speech Large Language Models} (SLLMs) empower the speech interaction with LLMs.
\textit{Continuous SLLMs}~\citep{wu2023decoder,chu2024qwen2audiotechnicalreport} adopt the speech encoders~\citep{radford2022robustspeechrecognitionlargescale} to extract continuous speech embeddings for LLM, which, however, only support speech understanding, relying on external TTS modules for speech generation, and therefore, hampering real-time interaction.
\textit{Discrete SLLMs}~\citep{zhang2023speechgpt}, instead, first discretize speech signals with speech tokenizers, followed by auto-regressive modeling.
Recent works~\citep{fang2024llamaomniseamlessspeechinteraction,xie2024mini} further combine the continuous speech encoders with discrete speech tokenizers for better results. 
Although effective, none of the existing works explore speech style controllability in SLLMs (\eg, emotions and pitches), which is essential for spoken dialogue.


\vspace{-2mm}
\paragraph{Omni-modal Large Language Models} support visual, text, and speech abilities with a unified architecture.
Similar to continuous SLLMs, InternOmni~\citep{chen2024internvlscalingvisionfoundation} and VITA~\citep{fu2024vitaopensourceinteractiveomni} connect a speech encoder with VLLMs, supporting speech understanding only.
Instead, AnyGPT~\citep{zhan2024anygptunifiedmultimodalllm} proposes a unified architecture to discretize and conduct auto-regressive modeling for image, text, and audio simultaneously, which, however, suffers from inevitable information loss brought by discretization, especially for the high-resolution visual inputs.
Our \textbf{\names} is the \textbf{very first} unified Omni-modal LLM with state-of-the-art vision-language and speech performance at the same time. 

%% file: section_arxiv/3_arch.tex
\section{Architecture}
\label{sec:method}


\subsection{Formulation}
\label{sec:notations}
Denote LLM as $f(\cdot)$ and text, visual and speech inputs as $\textbf{X}_T$, $\textbf{X}_V$ and $\textbf{X}_S$, respectively.
$\textbf{X}_T$ is converted to discrete tokens $\textbf{U}_T$ via a text tokenizer~\citep{gage1994new}, while the $\textbf{X}_V$ is first encoded with a vision encoder $v(\cdot)$ as $\textbf{E}_V = v(\textbf{X}_V)$, and then projected into the textual embedding space with a projector $p(\cdot)$ as $\textbf{H}_V = p(\textbf{E}_V)$.
As for the speech input $\textbf{X}_S$, a \textit{Speech-to-Unit} (S2U) procedure is required.
Specifically, $\textbf{X}_S$ first goes through a speech encoder $s(\cdot)$ as $\textbf{E}_S = s(\textbf{X}_S)$, which is then discretized by the quantizer $q(\cdot)$ as $\textbf{U}_S = q(\textbf{E}_S)$.
The LLM $f(\cdot)$ is then trained to compute the joint probability of the output texts $\textbf{U}_T^{o}$ and speech units $\textbf{U}_S^{o}$ as
\begin{equation}
    \mathbb{P}(\textbf{U}_T^{o},\textbf{U}_S^{o} | \textbf{U}_{omni}) = \prod_{i=1}^L \mathbb{P}(\boldsymbol{x}_i|\textbf{U}_{T,<i}^{o},\textbf{U}_{S,<i}^{o},\textbf{U}_{omni}),
    \label{equ:probability}
\end{equation}
where $\boldsymbol{x}_i\in \textbf{U}_T^{o}\cup\textbf{U}_S^{o}$, $L=|\textbf{U}_T^{o}|+|\textbf{U}_S^{o}|$ and
$\textbf{U}_{omni} = \textbf{U}_T\cup\textbf{U}_S\cup\textbf{H}_V$, which stands for the omni-modal inputs.
The output response units $\textbf{U}_S^{o}$ are then recovered into the output speech waveform $\textbf{Y}_S^{o}$ via a \textit{Unit-to-Speech} (U2S) decoder $d(\cdot,\cdot)$ with an emotion style embedding $\textbf{E}_{style}^o$ to realize the vivid emotional spoken dialogue controllability (Sec.~\ref{sec:speech_tokenizer}).


\paragraph{LLM.} 
We utilize the Qwen-2.5~\cite{qwen2.5} model families as the base LLMs of \textbf{\names} with three configurations (\ie, 3B, 7B, and 72B) for usage under different budgets.

\vspace{-2mm}
\paragraph{Vision encoder and projector.}
We use the QwenViT \cite{Qwen2-VL} as the visual encoder $v(\cdot)$ with an MLP vision projector $p(\cdot)$ with a $4\times$ downsample rate for all variants of \textbf{\names}.


\begin{figure*}[t]
\centering
\includegraphics[width=0.9\textwidth]{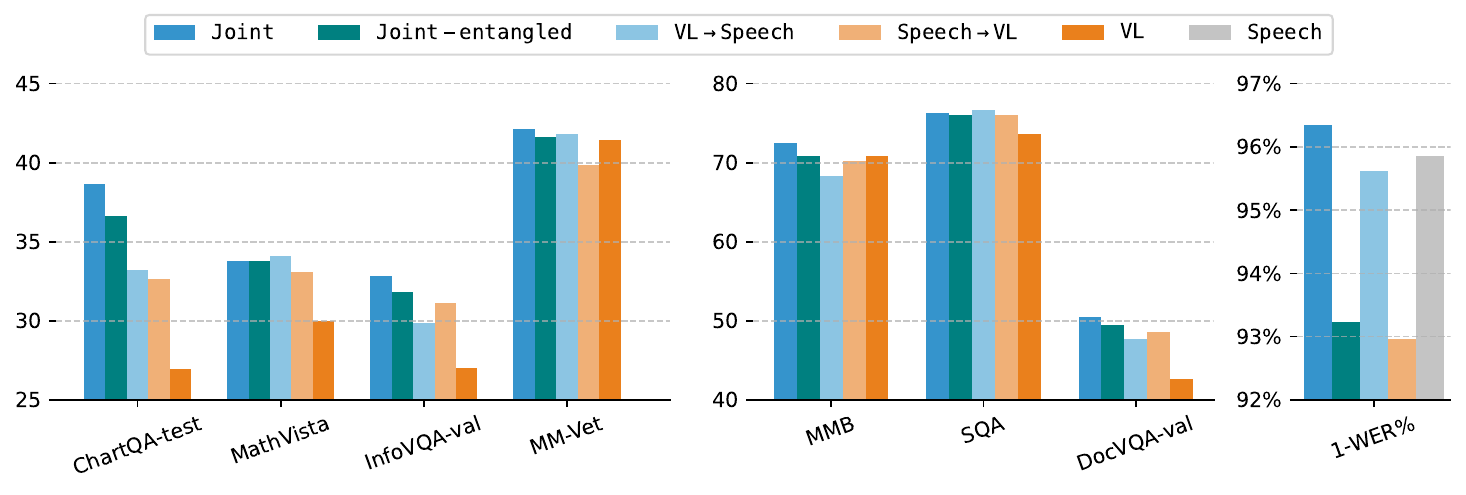}
\vspace{-4mm}
\caption{\textbf{Comparison between different omni-modal alignment paradigms.}
1) \texttt{Joint} training demonstrates consistent improvements over \texttt{VL} and \texttt{Speech}, suggesting that omni-modal alignment can be beneficial across modalities.
2) \texttt{Joint} training outperforms both \texttt{VL$\rightarrow$Speech} and \texttt{Speech$\rightarrow$VL}, revealing that joint training is more superior and efficient than sequential training.
3) \texttt{Joint} is superior to \texttt{Joint-entangled}, highlighting the effectiveness of the semantic-acoustic disentanglement, as discussed in Sec.~\ref{sec:speech_tokenizer}.
}
\vspace{-3mm}
\label{fig:speech_abla}
\end{figure*}


\subsection{Speech Tokenization}
\label{sec:speech_tokenizer}

\paragraph{Speech-to-unit (S2U) tokenizer.}
Following \cite{tao2024toneunitspeechdiscretizationapproach}, we use the SPIRAL~\citep{huang2022spiral} architecture for the speech encoder $s(\cdot)$ to capture both phonetic and tonal information, which is then discretized by the quantizer $q(\cdot)$ with finite scalar quantization (FSQ) \citep{mentzer2023finite}.
The size of the speech codebook is 4,096, while the sample rate is 25 tokens per second.
Once discretized, the speech modality can be integrated into LLMs by concatenating the text vocabulary and speech codebook.

Our S2U tokenizer provides the following advantages: 
1) \textit{Data efficiency}: after pre-training on large-scale unlabeled speech data, it requires only a small amount of speech-text pair data for easy adaptation.
2) \textit{Bilingual}: the speech codebook is shared among languages (\ie, English and Chinese), sharing unit modeling capabilities across languages.
Check more training details and comparisons in Appendix~\ref{app:s2u training}.


\vspace{-2mm}
\paragraph{Semantic-acoustic disentanglement.}
To align the speech units seamlessly with the highly semantic embedding space of LLMs, we opt for decoupling the semantic contents and acoustic styles of input speeches.
Given input speechs $\textbf{X}_S$, both semantic embedding $\textbf{E}_{semantic}$ and style embeddings $\textbf{E}_{style}$ are extracted separately as
\begin{equation}
    \{\textbf{E}_{semantic}, \textbf{E}_{style}\} = \textbf{E}_S = s(\textbf{X}_S).
    \label{equ:disentanglement}
\end{equation}
Only $\textbf{E}_{semantic}$ is quantified by $q(\cdot)$ to generate the speech units $\textbf{U}_S$.
By using different $\textbf{E}_{style}$ while maintaining the same $\textbf{E}_{semantic}$, we can easily control the recovered speech styles without disturbing the semantic contents of recovered speeches.
Moreover, the disentanglement facilitates modality alignment among speeches and texts, as later in Sec.~\ref{sec_speech_in}.


\vspace{-2mm}
\paragraph{Unit-to-speech (U2S) detokenizer with style controls.}
Building on VITS \citep{kim2021conditional}, our U2S detokenizer adopts a conditional VAE architecture (\cf, Fig.~\ref{fig:sdecoder}). 
To achieve flexible style controls, we utilize the semantic-acoustic disentanglement of our S2U tokenizer (as discussed above) and adopt a novel style embedding to control the speech styles (\eg, genders, emotions, and pitches).
Specifically, the LLM $f(\cdot)$ is trained to generate both the output speech units $\textbf{U}_S^{o}$ and a style label.
The speech units $\textbf{U}_S^{o}$ are converted to the unit embeddings $\textbf{E}_{semantic}^{o}$, while the style label is utilized to generate a unique style prototype $\textbf{E}_{style}^{o}$. 
Both $\textbf{E}_{semantic}^{o}$ and $\textbf{E}_{style}^{o}$ are taken as inputs to speech decoder $d(\cdot,\cdot)$ to synthesize output speeches $\textbf{Y}_{S}^{o} = d(\textbf{E}_{semantic}^{o},,\textbf{E}_{style}^{o})$.

Our U2S detokenizer is pre-trained on LibriTTS~\citep{zen2019libritts} and AISHELL-1~\citep{bu2017aishell} and subsequently fine-tuned on synthetic style-rich speech data.
Due to the scarcity of real-life style-rich data, we utilize TTS tools~\citep{du2024cosyvoice} to synthesize the speech samples diverse in genders, pitches, and emotions. 
As for the style prototypes, Emotion2Vec~\citep{ma2023emotion2vec} is adopted to select the most representative samples with the highest confidence in conveying the desired style.
Our empirical results reveal that even one representative style reference speech has been sufficient to control the speech styles flexibly and precisely.
Check Appendix~\ref{app: sec_u2s} for more details.

%% file: section_arxiv/4_alignment.tex

\begin{figure*}[t]
\centering
\includegraphics[width=0.8\textwidth]{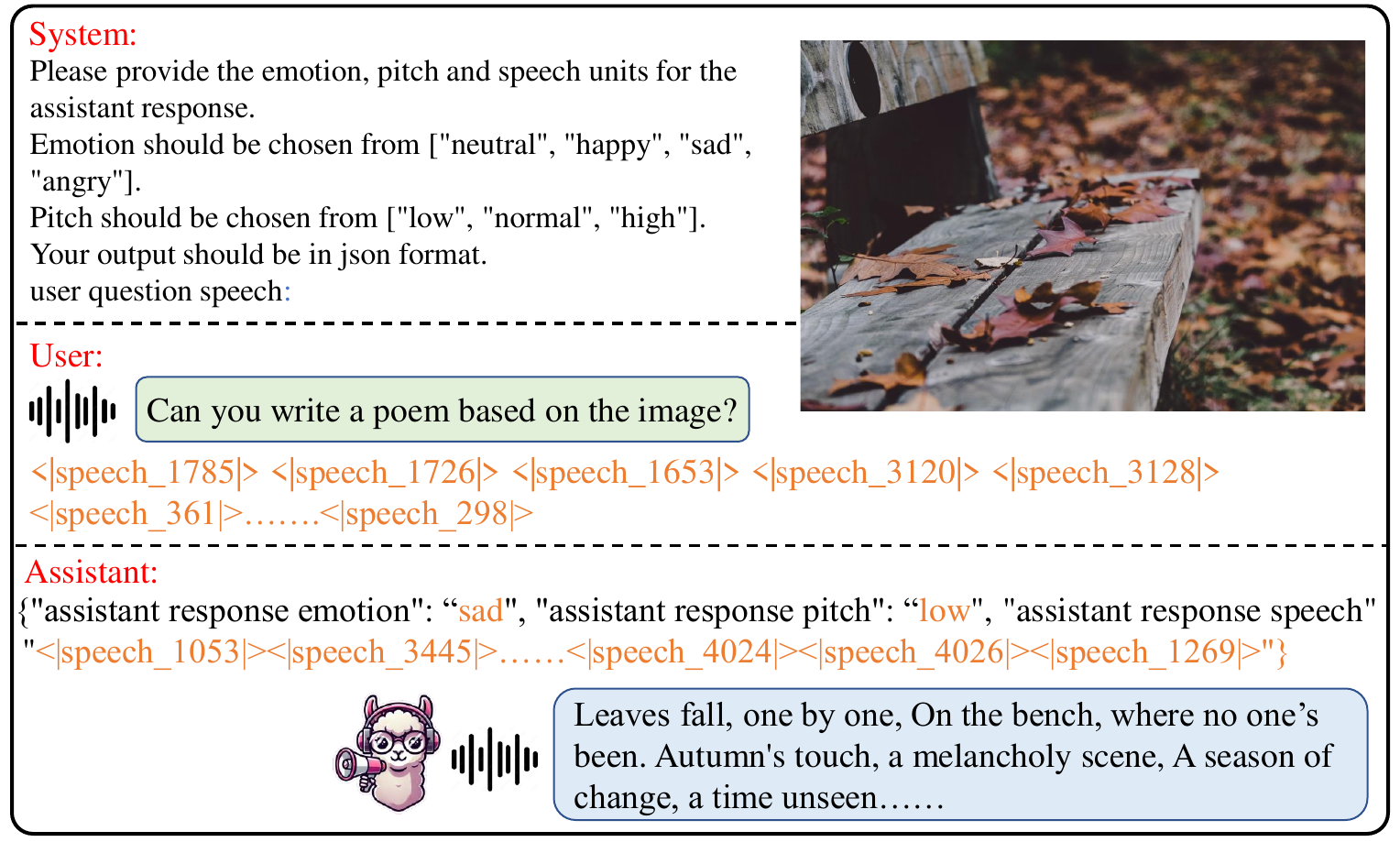}
\vspace{-3mm}
\caption{\textbf{Demonstration of \names omni-modal instruction tuning.}
1) To support emotional spoken dialogues, \textbf{\names} is trained to explicitly select speech style labels with output speech units. 
2) For ease of parsing, the data elements are organized in the \texttt{JSON} format.
}
\vspace{-2mm}
\label{fig:sft_prompt}
\end{figure*}


\section{Training Omni-modal LLMs}
\label{sec:omni_model_alignment}

To achieve omni-model alignment, it is ideal to use large-scale \textit{omni-modal image-text-speech} data, which, however, is either without reach due to copyrights~\citep{nagrani2022learning} or limited in quality~\citep{miech19howto100m}.
An alternative is to use the existing image-text data with the TTS-synthesized speeches, which is not only computationally expensive but also hampers data diversity, as most TTS tools generate speeches in similar patterns.
Recent works~\citep{chen2024internvlscalingvisionfoundation,fu2024vitaopensourceinteractiveomni} choose to integrate the speech modality into a well-structured VLLM via a \textit{sequential} training manner with \textit{bi-modal} alignment data.
\textit{However, the relationships among modalities and how to effectively leverage multiple bi-modal alignment datasets remain unclear. }

In this work, we explore omni-modal text-centric alignment by utilizing the publicly available bi-modal alignment datasets, including image-text (\eg, image captioning) and speech-text  (\eg, ASR and TTS) datasets.
With text modality as a bridge, our \textbf{\names} ultimately becomes a unified system capable of understanding and generating multiple modalities in a coherent and integrated manner. 
In Sec.~\ref{sec_speech_in}, we first explore the following three questions: 
\begin{enumerate}
    \item \textit{Does the integration of the speech modality conflict with the vision-language capabilities?}
    \item \textit{Is sequential alignment of multiple modalities optimal?}
    \item \textit{How to represent speech modality to foster omni-modal alignment?}
\end{enumerate}
We then discuss the omni-modal instruction tuning pipeline and the overall training paradigm of \textbf{\names} in Sec.~\ref{sec:ommi instruction tuning} and Sec.~\ref{sec_three_stage}, respectively.

\subsection{Omni-modal Text-Centric Alignment}
\label{sec_speech_in}

\paragraph{Settings.}
We consider the following omni-modal training paradigms:
1) \texttt{VL$\rightarrow$Speech} conducts the image-text alignment first followed by speech-unit-text alignment using the full speech data with 10\% of the image-text alignment data to prevent catastrophic forgetting, similarly with \citep{chen2024internvlscalingvisionfoundation,fu2024vitaopensourceinteractiveomni}.
2) \texttt{Speech$\rightarrow$VL} instead performs speech-unit-text alignment first and then aligns images with texts using 10\% of the speech unit-text data and full image-text data.
3) \texttt{Joint} aligns both modalities simultaneously. Note that unless otherwise specified, we utilize the S2U tokenizer introduced in Sec.~\ref{sec:speech_tokenizer} to extract speech units for all speech data, which effectively disentangles the semantic and acoustic features. 
4) \texttt{Joint-entangled} derives the speech units using HuBERT~\citep{10.1109/TASLP.2021.3122291}, which does not achieve semantic-acoustic disentanglement effectively with only K-means clustering.
5) \texttt{VL} and \texttt{Speech} only align vision and speech modalities with texts, respectively, serving as bi-modal baselines (see Appendix~\ref{app:text-centric alignment} for more details).

\vspace{-2mm}
\paragraph{Evaluation.} 
For speech abilities, we evaluate the aligned model's performance on the ASR task of LibriSpeech~\citep{Librispeech}, while for the vision-language, we fine-tune the model with a small amount of high-quality visual instruction data (\ie, 665K SFT data from the ShareGPT4V \citep{chen2023sharegpt4v}) and evaluate the fine-tuned model on common vision-language benchmarks. 
Check Appendix~\ref{sec_exp_vl_bench} for evaluation details.
Fig.~\ref{fig:speech_abla} shows the comparison among different paradigms on vision-language (left and middle) and ASR (right, where we report the $1-\text{WER}$ value for better readability) benchmarks, from which we can derive the following observations:

\vspace{-2mm}
\paragraph{Observation 1: image-text and speech-unit-text data benefit each other.}
Contrary to the common assumption that multiple modalities might compete and create conflicts, we notice that introducing additional modalities is actually beneficial. 
As in Fig.~\ref{fig:speech_abla}, \texttt{Joint} consistently surpasses both \texttt{VL} and \texttt{Speech} across vision-language and speech benchmarks. 
Moreover, even models aligned sequentially, (\ie, \texttt{VL$\rightarrow$Speech} and \texttt{Speech$\rightarrow$VL}, which are typically prone to catastrophic forgetting, demonstrate superior performance on most vision-language tasks. 
We speculate that the requirement to align multiple modalities with text leads to more robust representations, which in turn generalize better across different downstream tasks. 
This finding aligns with ImageBind~\citep{girdhar2023imagebind}, where joint alignment of audio and depth with images results in improved performance.

\vspace{-2mm}
\paragraph{Observation 2: semantic-acoustic disentanglement benefits omni-modal alignment.}
We find that 1) \texttt{Joint} outperforms \texttt{Joint-entangled} on vision-language benchmarks, and 2) in the speech tasks, \texttt{Joint} maintains significant advantages over its entangled counterpart. 
This can be attributed to the semantic-acoustic disentanglement which makes speech units more analogous to languages.

\input{table/main_table_vertical}

\vspace{-2mm}
\paragraph{Observation 3: sequential alignment is not optimal.}
We notice that \texttt{Joint} consistently outperforms its sequential counterparts (\ie, \texttt{VL$\rightarrow$Speech} and \texttt{Speech$\rightarrow$VL}) on both vision-language and speech benchmarks, probably due to catastrophic forgetting when integrating a new modality. 
In light of these observations, we have chosen to pursue the ultimate alignment strategy that simultaneously aligns image-text and speech-unit-text for \textbf{\names}, which offers two important benefits, 1) it fosters the mutual enhancement among vision-language and speech, and 2) it avoids catastrophic forgetting during sequential alignment.


\vspace{-1mm}
\subsection{Omni-modal Instruction Tuning}
\label{sec:ommi instruction tuning}
\vspace{-1mm}

After the omni-modal text-centric alignment in Sec.~\ref{sec_speech_in}, the model learns the fundamental vision-language (\eg, captioning) and speech capabilities (\eg, ASR and TTS).
However, instruction tuning is essential to better follow complicated user instructions and respond with vivid emotions. 

\vspace{-5mm}
\paragraph{Emotion-enriched instruction data synthesis.} 
Due to the scarcity of omni-modal instruction data (\ie, dialogues involving images, speeches, and texts simultaneously), we opt for synthesizing omni-modal instruction data from existing text and visual instruction datasets. 
First, we select instruction data suitable for the vocal expression by filtering out the non-vocal data (\eg, code and mathematical formulas). 
Second, we clean the selected data to be more vocal by removing text formatting elements (\eg, $**$ and \textbackslash n\textbackslash n).
We then obtain style labels for the remaining dialog contexts, including genders (\texttt{male, female}), pitches (\texttt{normal, low, high}),
and emotions (\texttt{happy, sad, angry, neutral}), resulting in totally $24$ different speech styles.
The style labels are generated by prompting GPT-4o
\footnote{https://chatgpt.ust.hk}
to make reasonable inferences given the dialogue context.
Finally, we convert the textual instructions and responses into speeches utilizing the latest TTS tools (\ie, CosyVoice~\citep{du2024cosyvoice} and Azure AI Speech), and the style labels are used to control the style of synthesized speech data. 
To further improve the diversity of the data, each instruction is synthesized by randomly choosing one of the 39 available speakers. 
Finally, we gather 120K speech-text and 110K speech-image data pairs. Check more details in Appendix~\ref{app:data synthesis}.

\vspace{-4mm}
\paragraph{Data organization and the chain of modality.}
The omni-modal instruction data can be represented as $D_\text{omni} = \{(x_V, u_S, x^o_T, c^o_\text{style}, u^o_S)_i\}_{i=1}^{N}$, where the input consists of the optional queried image $x_V$ and the speech units of the instruction $u_S$, while the output consists of the textual response $x^o_T$, the predicted speech style labels $c^o_\text{style}$, and the output speech unit $u^o_S$. 
Note that we train \textbf{\names} to explicitly select styles (\eg, emotions and pitches), which are utilized to determine the corresponding style embedding for the U2S detokenizer (Sec.~\ref{sec:speech_tokenizer}).
Furthermore, since directly generating speech responses is challenging, we decompose the speech response procedure into three primary steps: 
1) recognizing user instructions into texts; 
2) generating textual responses based on the recognized instructions; 
3) generating the style labels and response speech units based on the textual responses. 
For ease of parsing during deployment, the target outputs are formatted as \texttt{JSON}, as in Fig.~\ref{fig:sft_prompt}.

\begin{table*}[t]
\centering
\resizebox{0.9\textwidth}{!}{%
\begin{tabular}{@{}l|c|cc|cc|cc@{}}
\toprule
\multirow{2}{*}{\textbf{Datasets}} &
  \multicolumn{1}{c|}{\multirow{2}{*}{\textbf{End-to-end$\uparrow$}}} &
  \multicolumn{2}{c|}{\textbf{Text response}} &
  \multicolumn{2}{c|}{\textbf{Style Categorization}} &
  \multicolumn{2}{c}{\textbf{Recognition/Synthesis}} \\
\multicolumn{1}{c}{} &
  \multicolumn{1}{c}{} & Unit In &
  Text In &
  Emotion &
  Pitch &
  WER/CER$\downarrow$&
  TTS-WER/CER$\downarrow$\\
  \midrule
Speech-Image-EN & 7.45 & 7.56 & 7.95 & 82.50  & 97.70 & 2.40  & 3.20  \\
Speech-Text-EN  & 6.85 & 6.90 & 7.38 & 81.20 & 84.70 & 6.90  & 2.90  \\
Speech-Image-ZH & 6.48 & 7.02 & 6.82 & 77.60 & 95.90 & 1.70  & 12.00 \\
Speech-Text-ZH  & 5.25 & 5.58 & 6.60 & 80.90 & 93.20 & 10.70 & 12.20 \\
\bottomrule
\end{tabular}%
}
\vspace{-3mm}
\caption{\textbf{Evaluation of \names-7B on Speech Dialogue.}
By default, we evaluate on the corresponding test set of the evaluated datasets.
}
\vspace{-6mm}
\label{tab:speech_sft_result}
\end{table*}


\subsection{Overall Training Paradigm}
\label{sec_three_stage}

Inspired by~\cite{chen2023sharegpt4v}, a three-stage training paradigm is adopted,
\begin{itemize}
\item \textbf{Stage-1: Vision-language pre-alignment.} 
The purpose is to align the visual features into the embedding space of LLMs. Only the vision projector $p(\cdot)$ is trained.
\item \textbf{Stage-2: Omni-modal text-centric alignment.} 
This stage performs vision-language and speech-language alignment jointly. We train the LLM $f(\cdot)$, vision projector $p(\cdot)$, and the deeper half of vision encoder $v(\cdot)$ layers.
\item \textbf{Stage-3: Omni-modal instruction tuning.} 
We organize different datasets with various types of instructions to learn generalization across tasks, as detailed in Sec.~\ref{sec:implementation}.
\end{itemize}

%% file: table/main_table_vertical.tex
\definecolor{light-gray}{gray}{0.6}
\definecolor{front-color}{RGB}{232, 242, 255}
\newcommand{\FrontColor}[1]{{\cellcolor{front-color}{#1}}}

\begin{table*}[t]
    \centering
    \setlength{\tabcolsep}{3pt}
    \renewcommand{\arraystretch}{1.0}
    \resizebox{\textwidth}{!}{
    \begin{tabular}{lccc|ccc|cc|ccc}
        \toprule
        \multirow{2}{*}{\textbf{Benchmarks}} & \multicolumn{1}{l}{\textbf{EMOVA}} & \multicolumn{1}{l}{\textbf{EMOVA}} & \multicolumn{1}{l|}{\textbf{EMOVA}} & \multicolumn{1}{l}{\textbf{Gemini}} & \textbf{GPT-} & \textbf{GPT-} & \textbf{Whisper} & \textbf{Mini-} & \textbf{VITA} & \textbf{VITA} & \multicolumn{1}{c}{\textbf{Baichuan-}} \\
       & \multicolumn{1}{c}{\textbf{3B}} & \multicolumn{1}{c}{\textbf{7B}} & \multicolumn{1}{c|}{\textbf{72B}} & \textbf{Pro 1.5} & \multicolumn{1}{c}{\textbf{4V}} & \multicolumn{1}{c|}{\textbf{4o}} & \multicolumn{1}{c}{\textbf{Large}} & \multicolumn{1}{l|}{\textbf{Omni2}} & \multicolumn{1}{c}{\textbf{8x7B}} & \multicolumn{1}{c}{\textbf{1.5}} & \multicolumn{1}{l}{\textbf{Omni-7B}} \\
        \midrule
        MME & \FrontColor 2175 & \FrontColor 2317 & \FrontColor {\bf 2402} & - & 1927 & 2310 & - & - & 2097 & 2311 & 2187 \\
        MMBench & \FrontColor 79.2 & \FrontColor 83.0 & \FrontColor {\bf 86.4} & - & 75.0 & 83.4 & - & - & 71.8 & 76.6 & 76.2 \\
        SEED-Image & \FrontColor 74.9 & \FrontColor 75.5 & \FrontColor 76.6 & - & 71.6 & \textbf{77.1} & - & - & 72.6$^*$ & 74.2 & 74.1 \\
        MM-Vet  & \FrontColor 57.3  & \FrontColor 59.4  & \FrontColor 64.8 & - & \textbf{67.7} & - & - & - & 41.6 & 51.1 & 65.4 \\
        RealWorldQA & \FrontColor 62.6 & \FrontColor 67.5 & \FrontColor 71.0 & 68.7 & 61.4 & \textbf{75.4} & - & - & 59.0$^*$ & 66.8 & 62.6 \\
        \midrule
        TextVQA & \FrontColor 77.2 & \FrontColor 78.0 & \FrontColor {\bf 81.4} & 73.5 & 77.4 & - & - & - & 71.8$^*$ & 74.9 & 74.3   \\
        ChartQA & \FrontColor 81.5 & \FrontColor 84.9 & \FrontColor {\bf 88.7} & 81.3 & 78.5 & 85.7 & - & -  & 76.6$^*$ & 79.6 & 79.6 \\
        DocVQA~\scriptsize{(test)} & \FrontColor 93.5 & \FrontColor 94.2 & \FrontColor {\bf 95.9} & 86.5 & 88.4 & 92.8 & - & - & -  & - & -  \\
        InfoVQA~\scriptsize{(test)} & \FrontColor 71.2 & \FrontColor 75.1 & \FrontColor {\bf 83.2} & 72.7 & - & - & - & -  & - & - & -   \\
        OCRBench & \FrontColor 803 & \FrontColor 814 & \FrontColor {\bf 843} & - & 656 & 736 & - & -  & 678 & 752 & 700 \\
        \midrule
        AI2D & \FrontColor 78.6 & \FrontColor 81.7 & \FrontColor {\bf 85.8} & 80.3 & 78.2 & 84.6 & - & -  & 73.1 & 79.3 & -   \\
        ScienceQA-Img & \FrontColor 92.7 & \FrontColor 96.4 & \FrontColor {\bf 98.2} & - & 75.7 & - & - & - & - & - & -  \\
        MMMU & \FrontColor 45.8 & \FrontColor 49.8 & \FrontColor 59.7 & 58.5 & 56.8 & \textbf{69.2} & - & - & 47.3 & 52.1 &  47.3 \\
        \midrule
        MathVista & \FrontColor 62.6 & \FrontColor 65.5 & \FrontColor {\bf 69.9} & 52.1 & 49.9 & 63.8 & - & -  & 44.9 & 66.2 & 51.9  \\
        Mathverse & \FrontColor 31.4 & \FrontColor 40.9 & \FrontColor {\bf 50.0} & - & 33.6 & - & - & - & -  & - & -   \\
        \midrule
        Librispeech~\scriptsize{(WER$\downarrow$)} & \FrontColor 5.4 & \FrontColor 4.1 & \FrontColor {\bf 2.9} & - & - & - & 3.0 & 4.8 & 3.4 & 8.1 & - \\
        \bottomrule
     \end{tabular}
    }
    \vspace{-3mm}
    \caption{\textbf{Comparison on vision-language and speech benchmarks.}
    1) \textbf{\names} outperforms GPT-4o/4V and Gemini Pro 1.5 on 11 of the 15 vision-language benchmarks, providing a powerful open-sourced alternative.
    2) Meanwhile, our \textbf{\names} achieves state-of-the-art performance on Librispeech, surpassing its speech and omni-modal counterparts significantly.
    $^*$: reported by~\cite{li2024baichuanomni}.
    }
    \vspace{-5mm}
    \label{tab:main_table}
\end{table*}

%% file: section_arxiv/5_experiment.tex
\vspace{-1mm}
\section{Experiments}
\label{sec:experiment}
\vspace{-1mm}

\subsection{Training configuration}\label{sec:implementation}

\paragraph{Stage-1.} 
In this stage, we only train the parameters of the vision projector $p(\cdot)$ for vision-language pre-alignment with the LCS-558K dataset~\citep{liu2023llava}, with the high-resolution image-slicing strategy~\citep{liu2024llavanext} adopted.

\vspace{-3mm}
\paragraph{Stage-2.}
We assemble a unified dataset with 7.4M samples for both the image-text and speech-text alignment, as summarized in Fig.~\ref{fig:pt_data}.
Specifically, we utilize pre-training datasets from ShareGPT4V \citep{chen2023sharegpt4v}, ALLaVA~\citep{chen2024allava} (both the original English version and the Chinese version translated on our own), and ShareGPT-4o~\citep{sharegpt4o} for general perception, while for
the OCR capabilities, we leverage SynthDog~\citep{kim2022donut}, MMC-Alignment~\citep{liu2023mmc}, K12 Printing, and the UReader Text Reading subset~\citep{ye2023ureader}.
Moreover, we adopt the 2,000 hours of ASR and TTS data from LibriSpeech~\citep{Librispeech} and AISHELL-2 \citep{aishell2} for speech-text alignment, and to preserve the language capabilities of LLMs, we further incorporate the text-only data from Magpie Pro~\citep{xu2024magpie}. Check more details in Fig.~\ref{fig:pt_data}.

\vspace{-3mm}
\paragraph{Stage-3.} 
We collect the \names-SFT dataset consisting of 4.4M multi-task omni-modal samples (see Fig.~\ref{fig:sft_data}).
We start by gathering high-quality open-sourced visual instruction datasets, including ShareGPT4V \citep{chen2023sharegpt4v}, InternVL~\citep{chen2024internvlscalingvisionfoundation}, Meteor~\citep{lee2024meteor}, Idefics-2~\citep{laurenccon2024matters}, Cambrian~\citep{tong2024cambrian}, and LLaVA-Onevision~\citep{li2024llavaonevisioneasyvisualtask}, followed by quality checking, re-formatting all data samples with a unified template, and removing the duplicated data.
For speech, we include the training split of \textbf{\names} omni-model instruction data (\cf, Sec.~\ref{sec:ommi instruction tuning}), with 10\% of speech alignment datasets to maintain ASR and TTS performance.
We train with 128 Ascend 910B (64GB) NPUs in parallel (check more details in Table~\ref{tab:implementation_detail}).


\subsection{Comparison to SOTA Models}

Experimental results are provided in Table~\ref{tab:main_table}. 
We compare a wide range of state-of-the-art VLLMs, including Gemini Pro 1.5~\citep{reid2024gemini}, GPT-4V~\citep{openai2023gpt4v}, GPT-4o \citep{openai2024gpt4o}, together with the Speech LLM (\ie, Mini-Omni2 \citep{xie2024mini}) together with the ASR expert Whisper-Large~\cite{radford2022robustspeechrecognitionlargescale}, and the omni-modal LLMs (\ie, VITA-8x7B~\citep{fu2024vitaopensourceinteractiveomni}, VITA-1.5~\cite{fu2025vita} and Baichuan-Omni~\cite{li2024baichuanomni}).

\vspace{-3mm}
\paragraph{Comparison with SOTA VLLMs.} 
As an omni-modal model, \textbf{\names} obtains comparable performance with the state-of-the-art VLLMs on multiple vision-language benchmarks, while showing superior proficiency in solving math problems needing precise visual content interpretation. Our \textbf{\names-7B} surpasses GPT-4V by +7.3 on MathVerse, and our \textbf{\names-72B} exceeds GPT-4o by +6.1 on MathVista.
On \textbf{11 out of the 15} benchmarks, our \textbf{\names} outperforms both GPT-4o/4V and Gemini Pro 1.5, providing a powerful open-sourced alternative.

\vspace{-3mm}
\paragraph{Comparison with SOTA omni-modal LLMs.} 
Under the similar 7B capacity, \textbf{\names-7B} is 220 points higher than VITA on MME, surpassing VITA by 20.6\% on OCRBench (814 vs. 678).
Moreover, it surpasses Baichuan-Omni-7B, a more recent Omni-modal LLM, on nearly all the evaluated benchmarks, underscoring the effectiveness of our approach and potential to push the boundary of omni-modal abilities.

\begin{figure*}[t]
\centering
\includegraphics[width=0.9\textwidth]{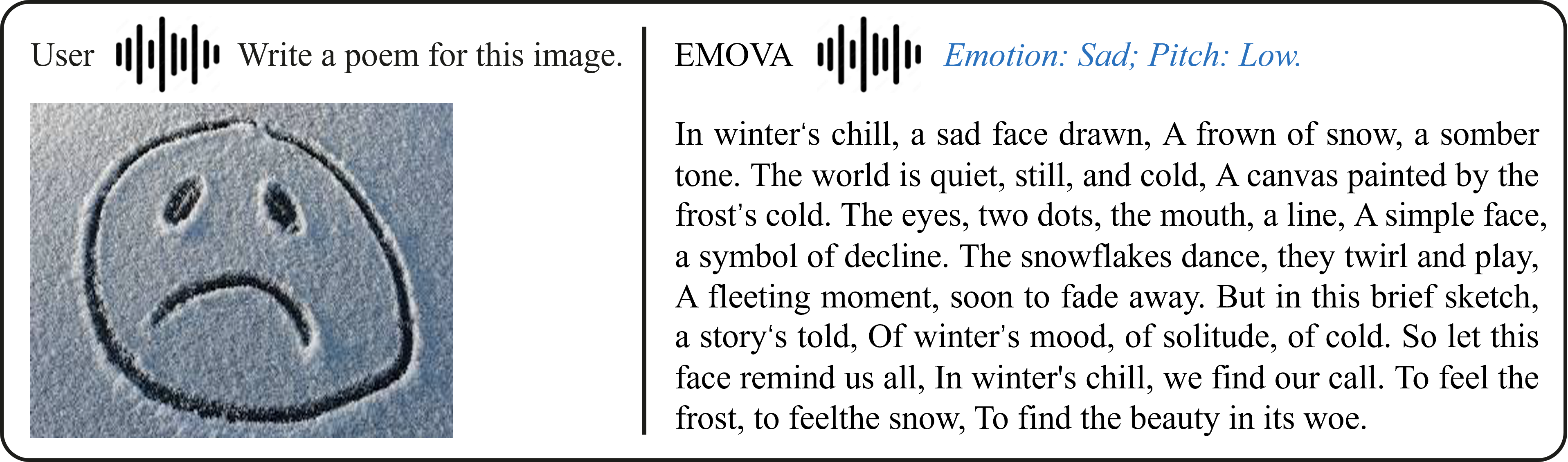}
\vspace{-3mm}
\caption{\textbf{\names} engages in omni-modal \textbf{emotional spoken dialogue} expressing sadness.}
\vspace{-3mm}
\label{fig:case_emo_poem}
\end{figure*}

Moreover, ASR results show that \textbf{\names} significantly outperforms the most recent omni-modal model VITA, even surpassing its SLLM counterparts Mini-Omni2 and ASR-specialized Whisper-Large, showing the effectiveness of the semantic-acoustic disentanglement and omni-modal mutual benefits.
We further report TTS results in Table~\ref{tbl_asr_tts}, and provide the qualitative examples in Fig.~\ref{fig:case_emo_poem} and Appendix~\ref{App:Qualitative Results}.
\textbf{For the first time}, \textbf{\names} obtains state-of-the-art results on both the speech and vision-language benchmarks.


\subsection{Evaluation of Emotion-Rich Spoken Dialogue}

In this section, we evaluate the end-to-end spoken dialogue abilities of \textbf{\names-7B}. As stated in Sec.~\ref{sec:ommi instruction tuning}, the model takes an input image $x_V$ and user instructions in the form of speech units $u_S$, and outputs \textit{text responses}, \textit{style labels}, and \textit{corresponding speech units}. To ensure comprehensive evaluation, we propose the following evaluation metrics:
\begin{enumerate}
    \item \textbf{End-to-end spoken dialogue score} assesses the model's dialogue performance based on the generated speeches, with a score ranging from 0 to 10, reporting the average.
    \item \textbf{Unit-input-text-output score} focuses on the quality of the textual responses of LLM when the inputs are speech units, bypassing errors from speech synthesis.
    \item \textbf{Text-input-text-output score} inputs the ground-truth user instruction texts and evaluates the model's text outputs. This helps disentangle the impact of speech recognition errors and eliminates the effect of \texttt{JSON} format.
    \item \textbf{ASR and TTS} evaluate how accurately the model recognizes the speech units and how effectively it generates speech units from text. See Appendix~\ref{app:speech evaluation} for more details.
    \item \textbf{Style label classification accuracy} evaluates the accuracy in selecting the appropriate speech style labels.
    \item \textbf{Style controllablity} assesses the controllability of U2S detokenizer with the given conditional style labels using the confusion matrix comparing the generated and recognized style labels. See Appendix~\ref{app:speech evaluation} for more details.
\end{enumerate}
Due to the lack of emotionally rich spoken dialogue evaluation datasets, we split a test set from our synthesized omni-modal instruction-tuning data (Sec.~\ref{sec_speech_in}). 
GPT-4o is used for automated evaluation. See details in Appendix \ref{app:speech evaluation}.

\vspace{-3mm}
\paragraph{Results.}
Table \ref{tab:speech_sft_result} the spoken dialogue performance.

\textbf{(i)} By comparing the \textit{end-to-end dialogue score} with the \textit{unit-input-text-output score}, we notice that the two scores are closely aligned, with a maximum gap of only $0.33$, except for Speech-Image-ZH. 
TTS-WER/CER is generally low for English, revealing that \textbf{\names} can synthesize accurate speech based on textual responses, which, however, is harder for Chinese, which we attribute to its complexity. 
It includes tasks such as generating poetries and answering riddles, resulting in more intricate responses. 


\begin{figure}[t]
\centering
\vspace{-2mm}
\includegraphics[width=0.9\columnwidth]{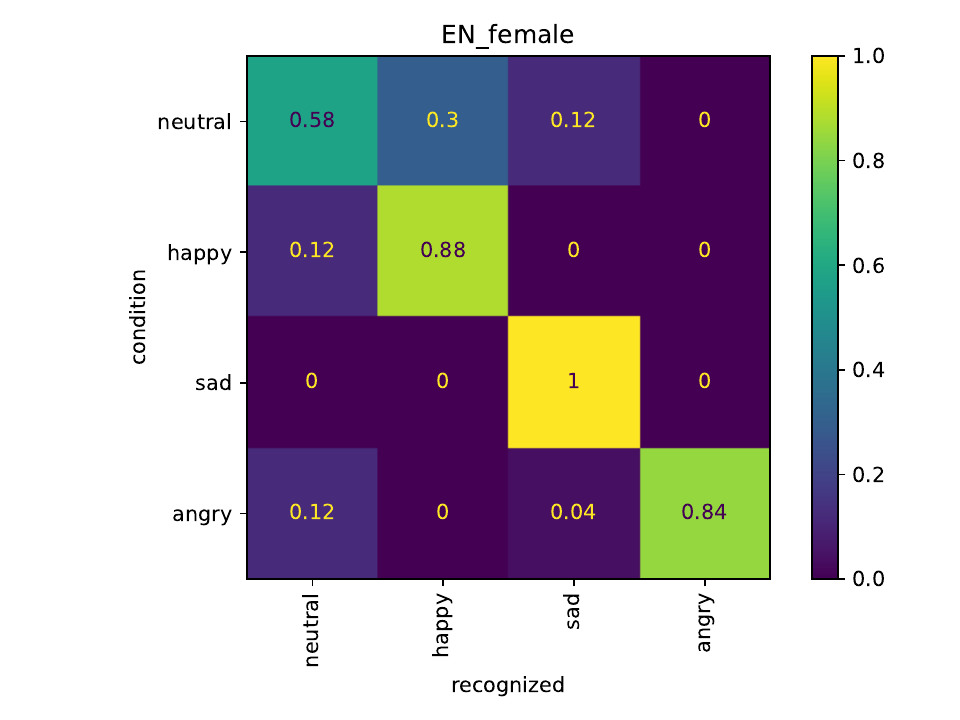}
\vspace{-4mm}
\caption{\textbf{Confusion matrix between the generated and recognized emotions.} The emotions generated by our U2S detokenizer are recognized with high probability. Best view with zooming in.}
\label{fig:confuison_matrix_EN_female}
\vspace{-4mm}
\end{figure}


\textbf{(ii)} Comparing the \textit{unit-input-text-output} score with the \textit{text-input-text-output} score, we notice that their differences correlate with the ASR performance of the speech instructions, especially for Speech-Text-EN and Speech-Text-ZH, which involve more complex instructions.

Our \textbf{\names-7B} reports inferior ASR performance ($6.9$ and $10.7$, respectively) compared to other datasets ($2.4$ and $1.7$). 
Consequently, when we replace speech instructions with ground-truth transcriptions, \textbf{\names} shows significant improvements from \textit{unit-input} to \textit{text input} score. 
On the contrary, for datasets with accurate ASR performance, the results are quite similar, suggesting \textbf{\names} retains robust dialogue capabilities when using the \texttt{JSON} format.

\textbf{(iii)} Examining the \textit{classification accuracy of style labels}, we find that \textbf{\names} performs satisfactorily in classifying emotions and pitches during speech conversations, achieving an accuracy of over 75\%. The confusion matrix comparing the conditional and recognized emotion labels is shown in Fig.~\ref{fig:confuison_matrix_EN_female}. The results indicate that the four emotions are recognized with high probabilities, with three achieving over 80\% accuracy. This demonstrates that our U2S detokenizer effectively controls common emotions, endowing the synthesized speech with vivid emotional expression.

%% file: section_arxiv/6_conclusion.tex
\section{Conclusion}

Our work builds \textbf{\names}, a novel end-to-end omni-modal large language model that effectively aligns vision, speech, and text simultaneously.
With text as a bridge, we show that omni-modal alignment is achievable without relying on omni-modal image-text-speech data, meanwhile, enhancing both vision-language and speech abilities.
For the first time, \textbf{\names} achieves state-of-the-art performance on both vision-language and speech benchmarks, setting a new standard for versatile omni-modal interactions.

%% file: section_arxiv/7_appendix.tex
\onecolumn
\appendix
\section*{Appendix} 

\section{More on Speech Tokenizer}
\subsection{Speech-to-Unit (S2U) Tokenizer}
\label{app:s2u training}

\paragraph{Overview.}
To process the speech input $\textbf{X}_S$, our S2U tokenizer consists of a speech encoder $s(\cdot)$ with a quantization module $q(\cdot)$. 
First, the speech input is passed through $s(\cdot)$, producing a continuous latent representation $\textbf{E}_S = s(\textbf{X}_S)$.
Then, the quantization module $q(\cdot)$ converts $\textbf{E}_s$ into discrete units $\textbf{U}_S = q(\textbf{E}_s)$. The final output is an ID sequence $\textbf{U}_S = [u_1, u_2, \cdots]$, where each $u_i$ corresponds to a unique speech unit in the speech codebook.

After this S2U extraction procedure, the speech is represented by quantized IDs instead of quantized embeddings. For example, a speech signal is represented as an ID sequence like $[782, 463, 550, \cdots]$, which can be treated as a special form of text. As a result, to integrate speech tokens into LLM $f(\cdot)$, we only need to expand the LLM's original vocabulary $V_T$ by adding a set of speech unit tokens $V_S$, similar to~\cite{zhang2023speechgpt}. The expanded vocabulary is thus the union $V = V_T \cup V_S$. 
In this work, the same codebook is shared across multiple languages, such as English and Chinese, enabling the unit modeling abilities to be shared across languages.
This design enables simply integration of the speech modality into the LLM with minimal data requirements (see experiments in Sec.~\ref{sec_speech_in}).

\paragraph{Training of S2U.}
The training of the S2U encoder involves three primary components: the speech encoder, the quantization module, and the phone decoder. 
First, the speech encoder is trained with a large amount of unlabeled speech with contrastive loss in a self-supervised learning manner~\citep{tao2024toneunitspeechdiscretizationapproach}. 
The dataset utilized is the 10000-hour English speeches from GigaSpeech~\citep{chen2021gigaspeech} and the 10000-hour Chinese speeches from the WenetSpeech~\citep{zhang2022wenetspeech}, both of which encode large variations in speakers, styles, and acoustic environments. 
Subsequently, the speech encoder, along with the quantization module and phone decoder, are optimized on a small amount of speech-text pair data, which is derived from the train-clean-100 subset of  LibriSpeech~\citep{Librispeech} in English and AISHELL-1~\citep{bu2017aishell} in Chinese. The phone label is obtained using the phone set in Charsiu~\citep{zhu2022phone}. During training, the speech encoder encodes input speeches into continuous latent representations that are rich in both phonetic and tonal information. Then, the quantization module is used to convert the continuous outputs from the speech encoder into discrete units. Finally, the phone decoder converts the quantized embeddings into a sequence of non-tonal/tonal phones, ensuring that the speech units capture necessary information related to semantic contents in both non-tonal and tonal languages.
After training, only the speech encoder and the quantization module are used in \textbf{\names}.

\paragraph{Comparisons with SpeechTokenizer in AnyGPT.}
Our S2U tokenizer differs from the SpeechTokenizer~\citep{zhang2023speechtokenizer} adopted in AnyGPT~\citep{zhan2024anygptunifiedmultimodalllm}, in the following aspects:

(1) SpeechTokenizer encodes both semantic contents and acoustic details of speeches, while our S2U tokenizer focuses solely on semantic contents. This design reduces the LLMs' burden of disentangling different aspects of speech information, facilitating the modality alignment between speech and text modalities during LLM training.

(2) Compared with SpeechTokenizer, our S2U tokenizer offers a more concise representation and helps to simplify and accelerate generation. SpeechTokenizer employs tokens from eight RVQ layers with a 50Hz frame rate to represent speech, thus a 10-second speech corresponds to 500 × 8 = 4000 tokens. However, we reduce the frame rate from 50Hz to 25Hz and utilize only one token to represent each frame, and thus, a 10-second speech can be represented by only 250 tokens. Moreover, AnyGPT requires a two-stage generation process, involving autoregressive (AR) semantic token generation followed by the non-autoregressive (NAR) acoustic token generation. Instead, we only need to generate speech units capturing the semantic contents in a fully AR manner.

(3) SpeechTokenizer lacks an explicit structure design to deal with tonal languages like Chinese, therefore, the processing ability in Chinese is not demonstrated in either SpeechTokenizer or AnyGPT. In contrast, our S2U tokenizer incorporates training constraints to better capture tone variation in phone, making it effective for both the non-tonal and tonal languages. This further enhances \textbf{\names}'s multilingual speech processing capabilities, enabling it to effectively handle both English and Chinese.

In summary, our S2U tokenizer improves the compactness and generality of speech representation, facilitates LLM training, and enhances its multilingual speech ability. Experimental results show that our model significantly outperforms AnyGPT in ASR tasks, as shown in Table~\ref{tbl_asr_tts}.


\begin{wrapfigure}{rt}{0.5\textwidth}
\centering
\vspace{-2mm}
\includegraphics[width=\linewidth]{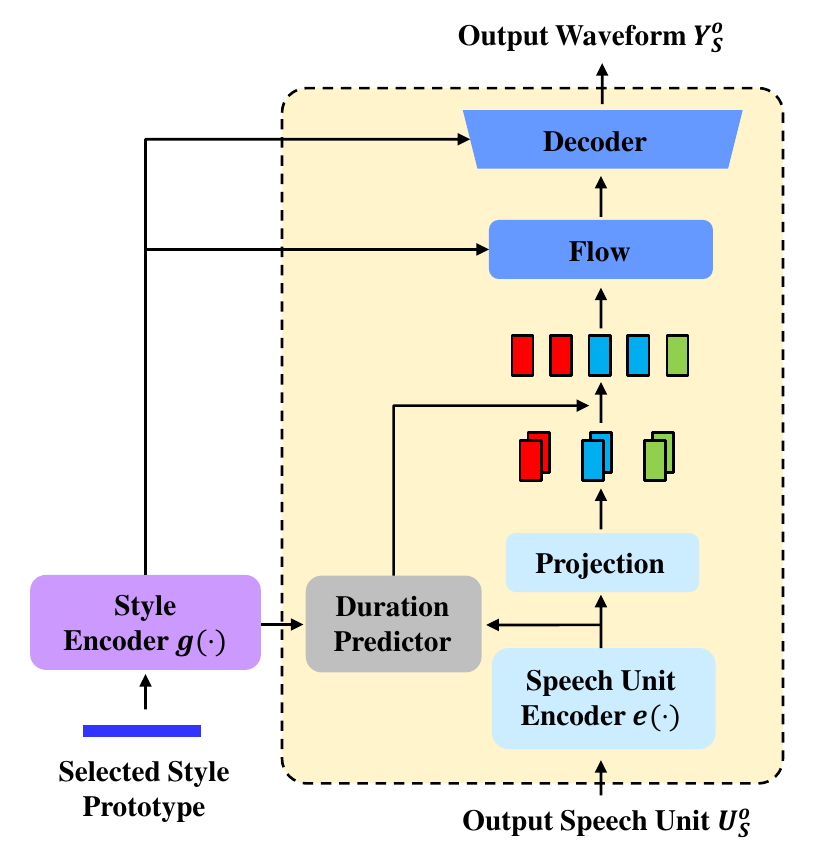}
\vspace{-3mm}
\caption{\textbf{U2S detokenizer with style control.}}
\vspace{-8mm}
\label{fig:sdecoder}
\end{wrapfigure}


\subsection{Unit-to-Speech (U2S) Detokenizer with Style Control}
\label{app: sec_u2s}

\paragraph{Overview.}
The LLM, along with the vision encoder and speech tokenizer, is trained end-to-end to generate responses in the form of the speech units, given the input images and speeches. Specifically, the output speech units can be obtained via $\textbf{U}_S^{o} = f(\textbf{U}_T,\textbf{U}_S,\textbf{H}_V)$, followed by a U2S detokenizer to convert the discrete speech units $\textbf{U}_S^{o}$ into the final output speech waveforms.

The proposed U2S detokenizer involves three core modules: the speech unit encoder $e(\cdot)$, the speech style encoder $g(\cdot)$, and the speech decoder $d(\cdot, \cdot)$. 
First, the speech unit encoder converts the speech units $\textbf{U}_S^{o}$ into unit embeddings $\textbf{E}_{unit}^{o}$.
Meanwhile, the style encoder $g(\cdot)$, adopting the structure of Meta-StyleSpeech~\citep{min2021meta}, is utilized to extract a style embedding $\textbf{E}_{style}^{o}$ from the chosen reference speech.
Lastly, the speech decoder $d(\cdot,\cdot)$ reconstructs the speech waveform $\textbf{Y}_{S}^{o}$ from the unit embedding $\textbf{E}_{unit}^{o}$ and style embedding $\textbf{E}_{style}^{o}$.

\paragraph{Training of U2S.}
Training a U2S detokenizer with emotion controls is challenging considering the lack of labeled emotional speech data since most open-source speech data is predominantly neutral in emotion or lacks emotion labels. 
Due to the limited availability of emotion-rich data, we utilize TTS tools~\citep{du2024cosyvoice} to generate a small set of style-rich speech samples diverse in speaker identities, genders, emotions, and pitches. 
Our U2S detokenizer is first pre-trained on LibriTTS~\citep{zen2019libritts} and AISHELL-1~\citep{bu2017aishell} to acquire fundamental speech synthesis capabilities, and subsequently, the synthesized style-rich speech data is utilized to fine-tune the U2S detokenizer, enhancing its controllability over diverse speech styles.

\paragraph{Style Prototypes.}
To better facilitate controls of genders, emotions, and pitches, inspired by \cite{min2021meta} that a small number of style reference speeches can effectively transfer the target styles, we adopt a ``store-for-usage" manner, \ie, we construct a style prototype codebook in advance for speech style assignation. 
Specifically, we synthesize $K$ reference candidates with external TTS tools for each possible combination of the following styles: two genders \texttt{(male, female)}, four emotions \texttt{(neutral, happy, sad, angry)}, and three pitches \texttt{(normal, high, low)}, leading to 24 unique styles and $24 \times K$ candidates. Empirically we find that genders and pitches are easy to control using any of the candidate references, while the emotion intensity varies across speeches. 
To tackle this, we adopt Emotion2Vec~\citep{ma2023emotion2vec}, a powerful speech emotion recognition (SER) tool, to measure the emotion intensity of each candidate reference, and rank them in terms of the confidence of the desired emotion. We select the Top-1 candidate reference in each combination style to be the prototype of this condition. Finally, the most representative 24 reference speeches are selected from the $24 \times K$ candidates.



\section{More on Omni-modality}
\subsection{Omni-modal Text-Centric Alignment}
\label{app:text-centric alignment}

\begin{figure}[t]
\centering
\includegraphics[width=\columnwidth]{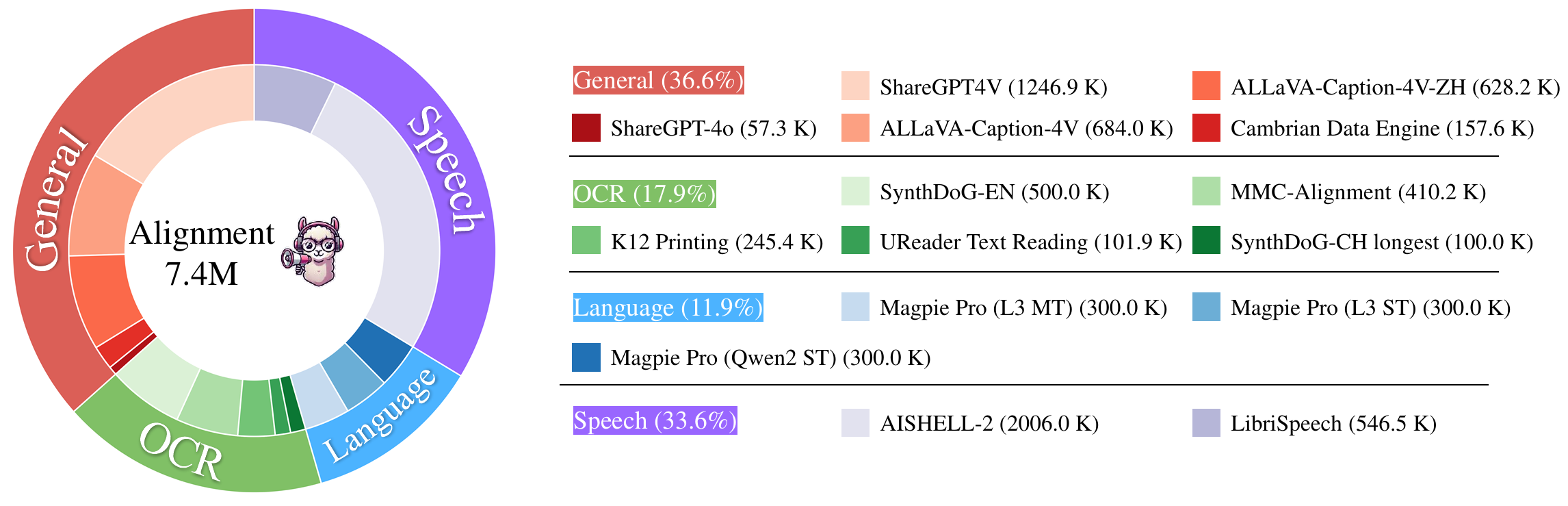}
\vspace{-3mm}
\caption{\textbf{Overview of \names omni-modal alignment data composition.}}
\vspace{-3mm}
\label{fig:pt_data}
\end{figure}

\paragraph{Modality alignment data} is summarized in Fig.~\ref{fig:pt_data}. 

\paragraph{Experiments on Omni-modal Alignment Paradigms.}
The training configuration adopted in Sec. \ref{sec_speech_in} is mostly identical to Table~\ref{tab:implementation_detail} except we use a unique resolution of 448 for all stages and replace EMOVA-SFT-4M in Stage-3 with ShareGPT4V~\citep{chen2023sharegpt4v} for efficiency. 

Given the space constraints, the evaluation benchmarks in Fig. \ref{fig:speech_abla} represent selected benchmarks from each category in Table \ref{tab:main_table}. Specifically, for general image perception and understanding, we choose MMBench and MM-Vet; for mathematical problem solving, we adopt MathVista (testmini); for science understanding, we select ScienceQA-Img; and for automatic speech recognition (ASR), we utilize the test-clean split of the LibriSpeech dataset.


\subsection{Omni-modal Instruction Data Synthesis}
\label{app:data synthesis}

\begin{figure}[t]
\centering
\includegraphics[width=\columnwidth]{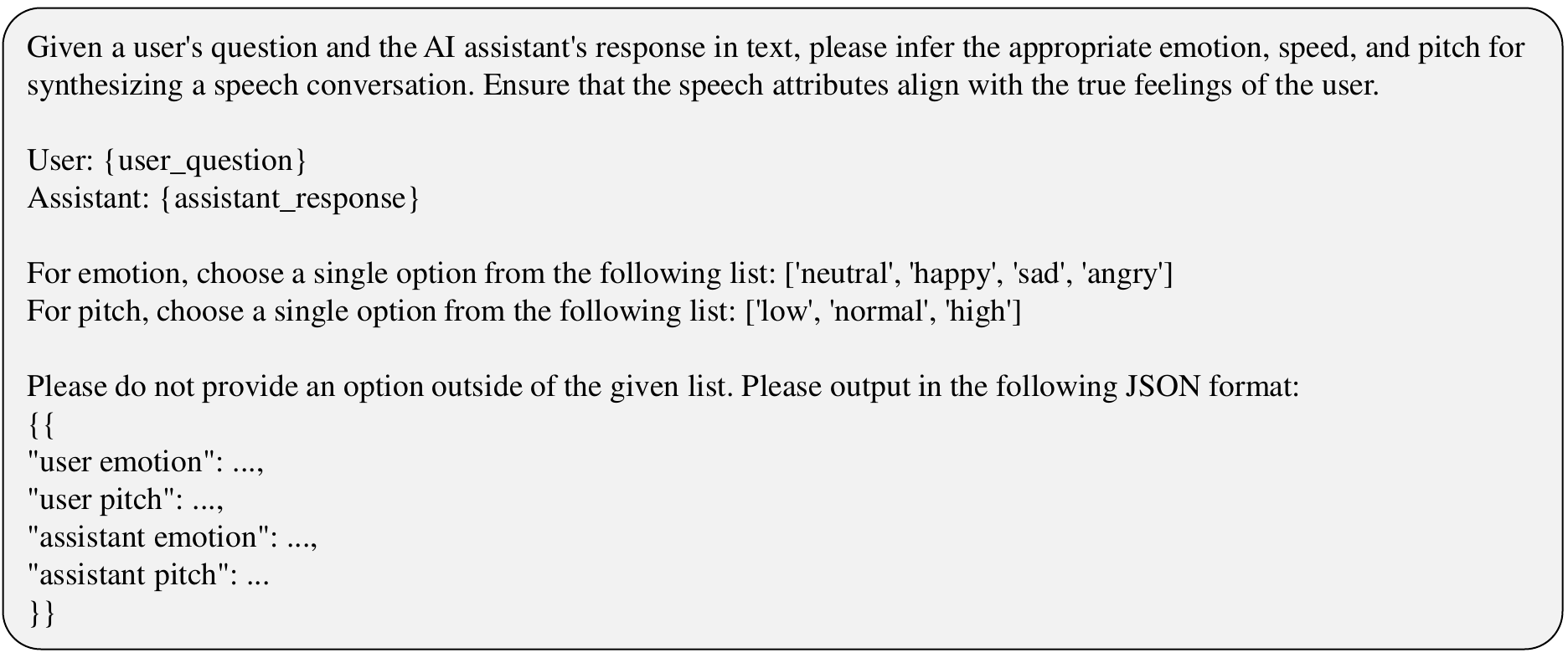}
\caption{\textbf{Prompt} used to obtain \textbf{style labels of the speech instruction dataset}.}
\label{fig:4o_style_label}
\end{figure}

\paragraph{Dataset construction.}
To obtain emotion and pitch labels, we leverage GPT-4o using the prompt in Fig. \ref{fig:4o_style_label}. 
Table~\ref{tab:emo_pitch_stat} shows the distribution of speech styles of our speech instruction dataset. 


\begin{figure}[t]
\centering
\includegraphics[width=\columnwidth]{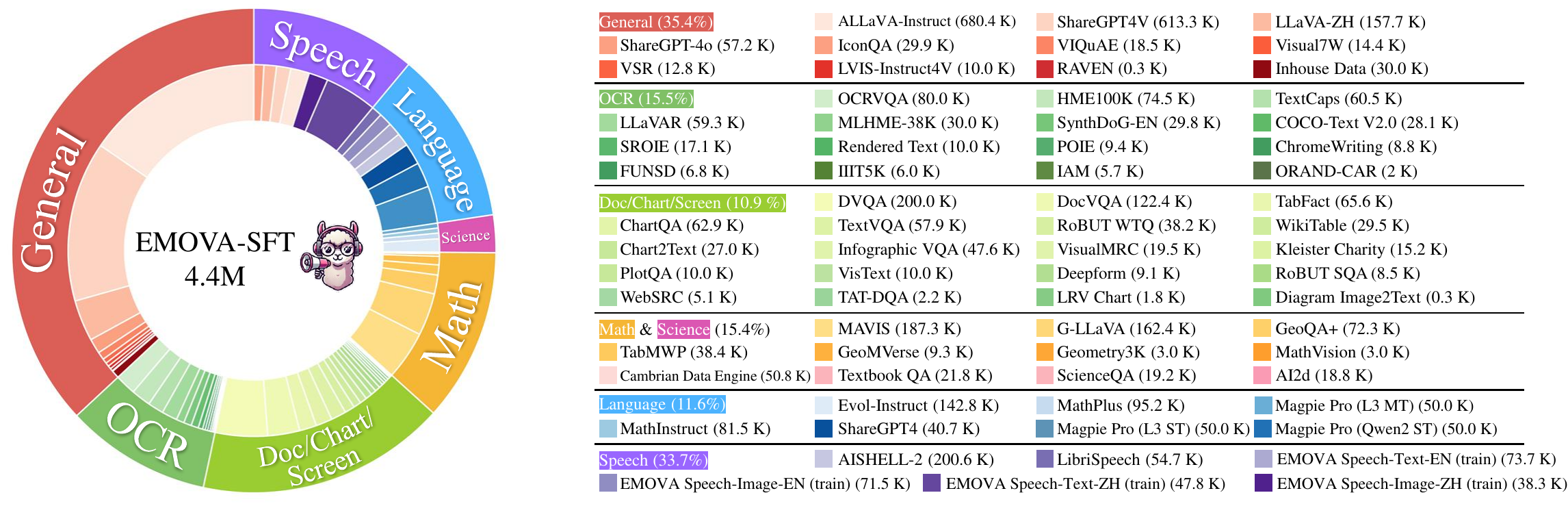}
\caption{\textbf{Overview of the data composition for \names omni-modal instruction tuning.} 
(Left) Distribution of instruction data across categories, with the outer circle representing overall categories and the inner circle depicting subset distributions. 
(Right) Quantitative breakdown of data sources.}
\label{fig:sft_data}
\end{figure}


\paragraph{Detailed data organization.}
As previously discussed in Sec. \ref{sec:ommi instruction tuning}, the omni-modal instruction data is formulated as $D_\text{omni} = \{(x_V, u_S, x^o_T, c^o_\text{style}, u^o_S)_i\}_{i=1}^{N}$. In details, the textual outputs $x_T^{o} = (x_T^{o^1}, x_T^{o^2})$ contain the transcribed textual instructions $x_T^{o^1}$ and the textual responses $x_T^{o^2}$. 
The styles labels $c^o_\text{style} = (c^o_\text{emo}, c^o_\text{p})$ include the emotion and pitch labels, respectively. 

\paragraph{Mathematical formulation of chain of modality.}
Based on the notations above, the sequential chain of modality approach can be mathematically formulated by decomposing the conditional likelihood of the desired outputs $(x_T^{o^1}, x_T^{o^2}, c^o_\text{emo}, c^o_\text{p}, u^o_S)$ given the inputs $(x_V, u_S)$. Specifically, let $z_1 = x_T^{o^1}, z_2 = x_T^{o^2}, z_3 = c^o_\text{emo}, z_4 = c^o_\text{p}, \text{and}, z_5 = u^o_S$, the decomposition is expressed as:
\begin{equation}
    \mathbb{P}(x_T^{o^1}, x_T^{o^2}, c^o_\text{emo}, c^o_\text{p}, u^o_S \mid x_V, u_S) = \prod_{i=1}^5 \mathbb{P}(z_i \mid z_{1:i-1}, x_V, u_S).
\end{equation}


\begin{table}[t]
\centering
\resizebox{0.95\textwidth}{!}{%
\begin{tabular}{@{}llc|cccc|ccc@{}}
\toprule
\multirow{2}{*}{Dataset} &
  \multicolumn{1}{c}{\multirow{2}{*}{Source}} &
  \multirow{2}{*}{\# Examples} &
  \multicolumn{4}{c|}{Emotions} &
  \multicolumn{3}{c}{Pitches} \\
 &
  \multicolumn{1}{c}{} &
   &
  \multicolumn{1}{c}{Neutral} &
  \multicolumn{1}{c}{Happy} &
  \multicolumn{1}{c}{Sad} &
  \multicolumn{1}{c}{Angry} &
  \multicolumn{1}{c}{Normal} &
  \multicolumn{1}{c}{Low} &
  \multicolumn{1}{c}{High} \\
  \midrule
\begin{tabular}[c]{@{}l@{}}Speech-Image\\ -EN(train)\end{tabular} &
  ALLAVA &
  71,474 &
  58,506 &
  12,412 &
  516 &
  40 &
  70,962 &
  392 &
  120 \\
\begin{tabular}[c]{@{}l@{}}Speech-Image\\ -EN(test)\end{tabular} &
  ALLAVA &
  1,056 &
  434 &
  274 &
  300 &
  48 &
  44 &
  176 &
  16 \\
\begin{tabular}[c]{@{}l@{}}Speech-Image\\ -ZH(train)\end{tabular} &
  \begin{tabular}[c]{@{}l@{}}ALLAVA\\ (ZH)\end{tabular} &
  38,260 &
  29,893 &
  7,680 &
  607 &
  80 &
  36,363 &
  624 &
  1,273 \\
\begin{tabular}[c]{@{}l@{}}Speech-Image\\ -ZH(test)\end{tabular} &
  \begin{tabular}[c]{@{}l@{}}ALLAVA\\ (ZH)\end{tabular} &
  616 &
  96 &
  193 &
  190 &
  137 &
  381 &
  177 &
  58 \\
\midrule
\begin{tabular}[c]{@{}l@{}}Speech-Text\\ -EN(train)\end{tabular} &
  ShareGPT &
  73,658 &
  42,334 &
  20,946 &
  4,674 &
  5,704 &
  60,352 &
  5,518 &
  7,788 \\
\begin{tabular}[c]{@{}l@{}}Speech-Text\\ -EN(test)\end{tabular} &
  ShareGPT &
  1,400 &
  200	&
  400 &
  400	&
  400	&
  582	&
  422	&
  422 \\
\begin{tabular}[c]{@{}l@{}}Speech-Text\\ -ZH(train)\end{tabular} &
  In-house &
  47,852 &
  29,769 & 
  16,405 &
  1,362 &
  316 &
  42,356 &
  4,379 &
  1,117 \\
\begin{tabular}[c]{@{}l@{}}Speech-Text\\ -ZH(test)\end{tabular} &
  In-house &
  684 &
  96 &
  196 &
  196 &
  196 &
  458	&
  134 &
  92 \\
  \bottomrule
\end{tabular}%
}
\vspace{-3mm}
\caption{\textbf{Statistics of the \names speech instruction tuning datasets.}}
\label{tab:emo_pitch_stat}
\end{table}


\begin{table}[t]
  \centering
  \renewcommand{\arraystretch}{1.2}
  \resizebox{0.95\textwidth}{!}{%
  \begin{tabular}{ll|c|c|c}
  \toprule
  & \textbf{Settings} & \textbf{Stage-1} & \textbf{Stage-2} & \textbf{Stage-3} \\
  \midrule
  \multirow{2}{*}{\rotatebox[origin=c]{90}{\footnotesize \textit{Vision}}} & \textbf{Resolution} & Min $56^2$, Max $1792^2$   & Min $56^2$, Max $1792^2$ & Min $56^2$, Max $1792^2$ \\
  & \# Tokens & Min $4$, Max $4096$ & Min $4$, Max $4096$   & Min $4$, Max $4096$  \\
  \midrule
  \multirow{2}{*}{\rotatebox[origin=c]{90}{\footnotesize \textit{Data}}} & \textbf{Dataset} & LCS & \names-Alignment (Fig.~\ref{fig:pt_data}) & \names-SFT (Fig.~\ref{fig:sft_data}) \\
  & \# Samples & 558K & 7.4M & 4.4M \\
  \midrule
  \multirow{5}{*}{\rotatebox[origin=c]{90}{\footnotesize \textit{Training}}} & \textbf{Trainable} & Projector & Full Model (Half ViT) & Full Model \\
  & \textbf{Batch Size} & 256 & 256 & 128 \\
  & \textbf{Context Length} & 8192 & 8192 & 8192 \\
  & \textbf{LR: $p(\cdot)$} & $1\times 10^{-3}$ & $2\times 10^{-5}$ & $2\times 10^{-5}$ \\
  & \textbf{LR: $v(\cdot)$} & - & $2\times 10^{-5}$ & $2\times 10^{-6}$ \\
  & \textbf{LR: $f(\cdot)$}  & - & $2\times 10^{-5}$ & $2\times 10^{-5}$ \\
  & \textbf{Epoch} & 1 & 1 & 1 \\
  \bottomrule
  \end{tabular}
  }%
  \vspace{-1mm}
  \caption{\textbf{Detailed configuration for different training stages of \names.}
  The table illustrates the vision configurations, dataset characteristics, and training hyperparameters.
  }
  \vspace{-2mm}
  \label{tab:implementation_detail}%
\end{table}%


\section{More on Benchmark Evaluation}
\label{sec_exp_vl_bench}

To thoroughly evaluate our model's vision-language abilities, 15 benchmarks covering four different aspects of the real-life scenarios are utilized for a comprehensive assessment across multiple domains.
Moreover, Automatic Speech Recognition (ASR) and Text-to-speech (TTS) are adopted to evaluate speech-language abilities.

\vspace{-2mm}
\paragraph{Document/chart understanding and OCR abilities.}
Benchmarks including TextVQA~\citep{singh2019textvqa}, ChartQA \citep{masry2022chartqa}, DocVQA~\citep{mathew2021docvqa}, InfoVQA~\citep{mathew2022infographicvqa}, and OCRBench~\citep{liu2023ocrbench}, focus on recognition and understanding of structured data  (\eg, charts, documents, and characters), challenging the model to extract, comprehend, and reason with structural and textual data.
We adopt the corresponding test set for evaluation.

\vspace{-2mm}
\paragraph{General image perception and understanding.} MME~\citep{fu2024mmecomprehensiveevaluationbenchmark}, MMBench~\citep{liu2023mmbench}, SEED-Bench-Image~\citep{li2023seed}, MM-Vet~\citep{yu2023mmvet}, and RealWorldQA~\citep{grok} evaluate models on perception and reasoning among general vision domains, 
providing a comprehensive measurement of models' real-world generalization abilities.

\vspace{-2mm}
\paragraph{Mathematical problem solving.}
MathVista (testmini)~\citep{lu2024mathvista} and MathVerse (vision-mini)~\citep{zhang2024mathverse} evaluate the model’s ability to deal with diverse mathematical problems, including both arithmetic and complex reasoning questions across different levels of complexity.

\vspace{-2mm}
\paragraph{Science understanding.} 
MMMU~\cite{yue2024mmmu}, ScienceQA-Img~\citep{lu2022scienceqa} and AI2D~\citep{kembhavi2016ai2d} are used to assess models’ ability to deal with scientific questions and diagrams, which requires reasoning across various subjects and interpreting structured visual content.

\vspace{-2mm}
\paragraph{Automatic speech recognition (ASR).}
We utilize the test-clean set of LibriSpeech \citep{Librispeech} for English, reporting the Word Error Rate (WER) as the evaluation metric. For Chinese, evaluation is conducted on the test set of AISHELL-2 \citep{aishell2}, using the Character Error Rate (CER). Both WER and CER assess ASR performance, calculated by comparing the recognized texts with the ground-truth transcripts. 

\vspace{-2mm}
\paragraph{Text-to-speech (TTS).}
To evaluate the TTS abilities, we first prompt \textbf{\names} to generate speech units, which are then converted to the speech waveforms by our U2S detokenizer. Using the synthesized speech as input, we conduct ASR with Whisper-large-v3 and Paraformer-zh for English and Chinese, respectively, to obtain transcribed texts. We then compute the WER and CER between the ground truth texts and the transcribed texts as metrics for TTS. The resulting metrics are denoted as TTS-WER and TTS-CER for English and Chinese.


\section{More on Evaluation of Speech-Language Capabilities}
\label{app:speech evaluation}

\subsection{Calculation of Evaluation Metrics}

\begin{figure}[t]
\centering
\includegraphics[width=\columnwidth]{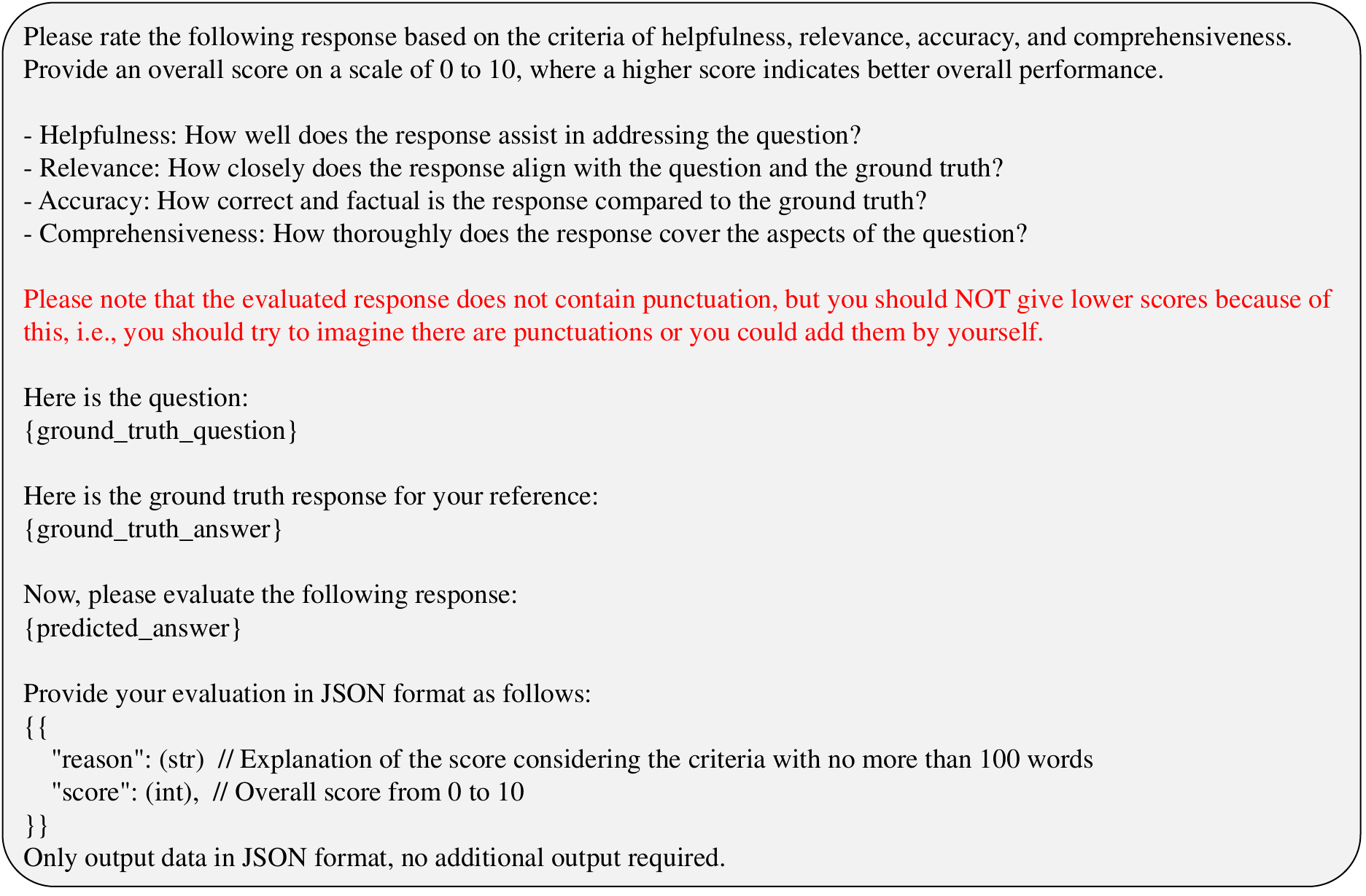}
\caption{\textbf{Prompt} used to obtain \textbf{End-to-end Spoken Dialogue Score}.}
\label{fig:4o_end2end}
\end{figure}

\paragraph{End-to-end spoken dialogue score.} We prompt GPT-4o with the original question $x_T^{o^1}$, the ground-truth text answer $x_T^{o^2}$ and the transcribed text from the generated speech, to obtain a score ranging from 0 to 10 and report an average of them. The prompt can be found in Fig. \ref{fig:4o_end2end}.

\begin{figure}[t]
\centering
\includegraphics[width=\columnwidth]{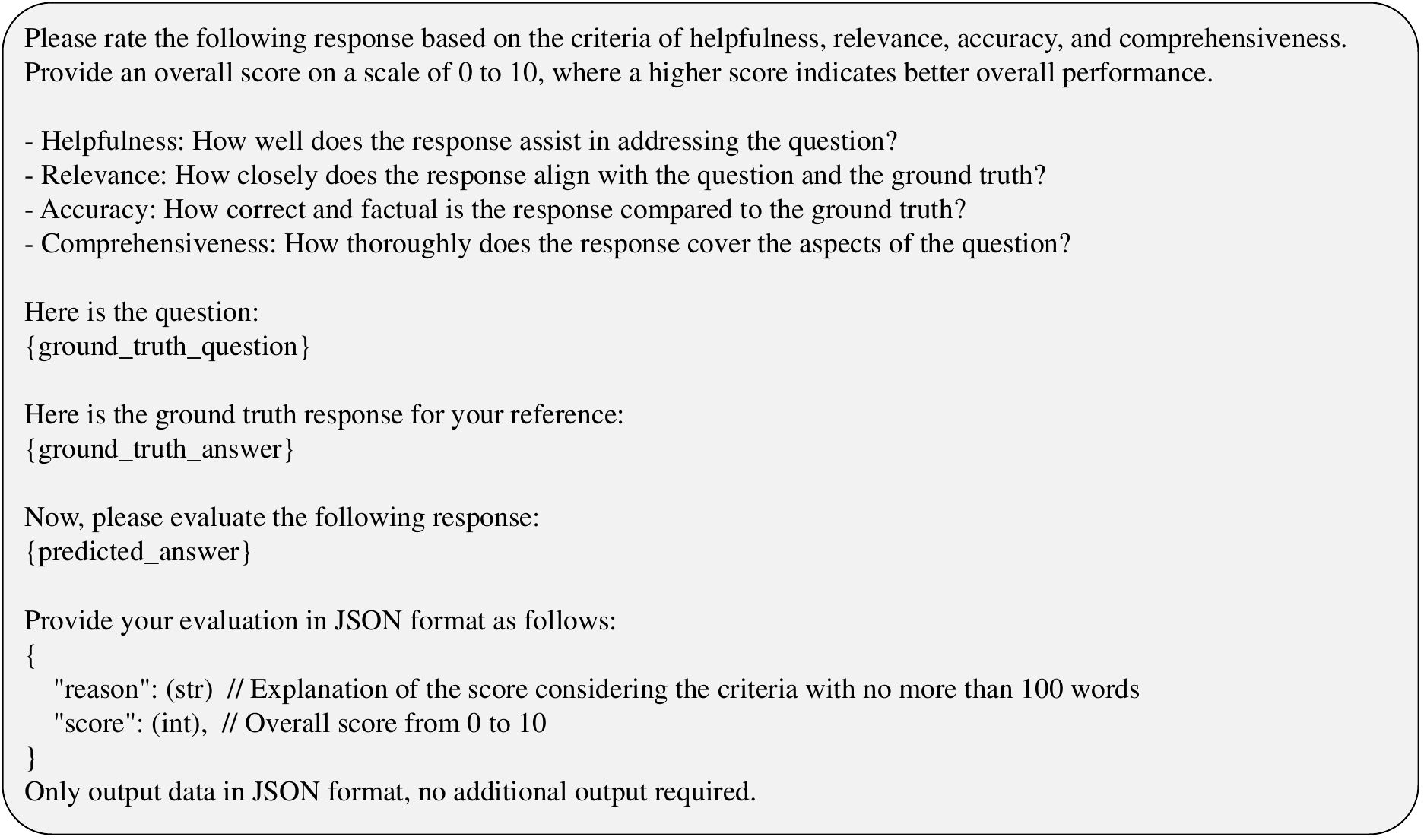}
\caption{\textbf{Prompt} used to obtain \textbf{Unit-Input-Text-Output Score} and \textbf{Text-Input-Text-Output Score}.}
\label{fig:4o_text}
\end{figure}

\paragraph{Unit-input-text-output score.} Similar to end-to-end spoken dialogue score, but we use the predicted text response $\tilde{x_T^{o^2}}$ as answer instead of the transcribed text from the generated speech, to obtain a score ranging from 0 to 10 and report an average of them. See the prompt in Fig.~\ref{fig:4o_text}.

\paragraph{Text-input-text-output score.} The prompt can be found in Fig. \ref{fig:4o_text}.

\begin{figure}[t]
\centering
\includegraphics[width=\columnwidth]{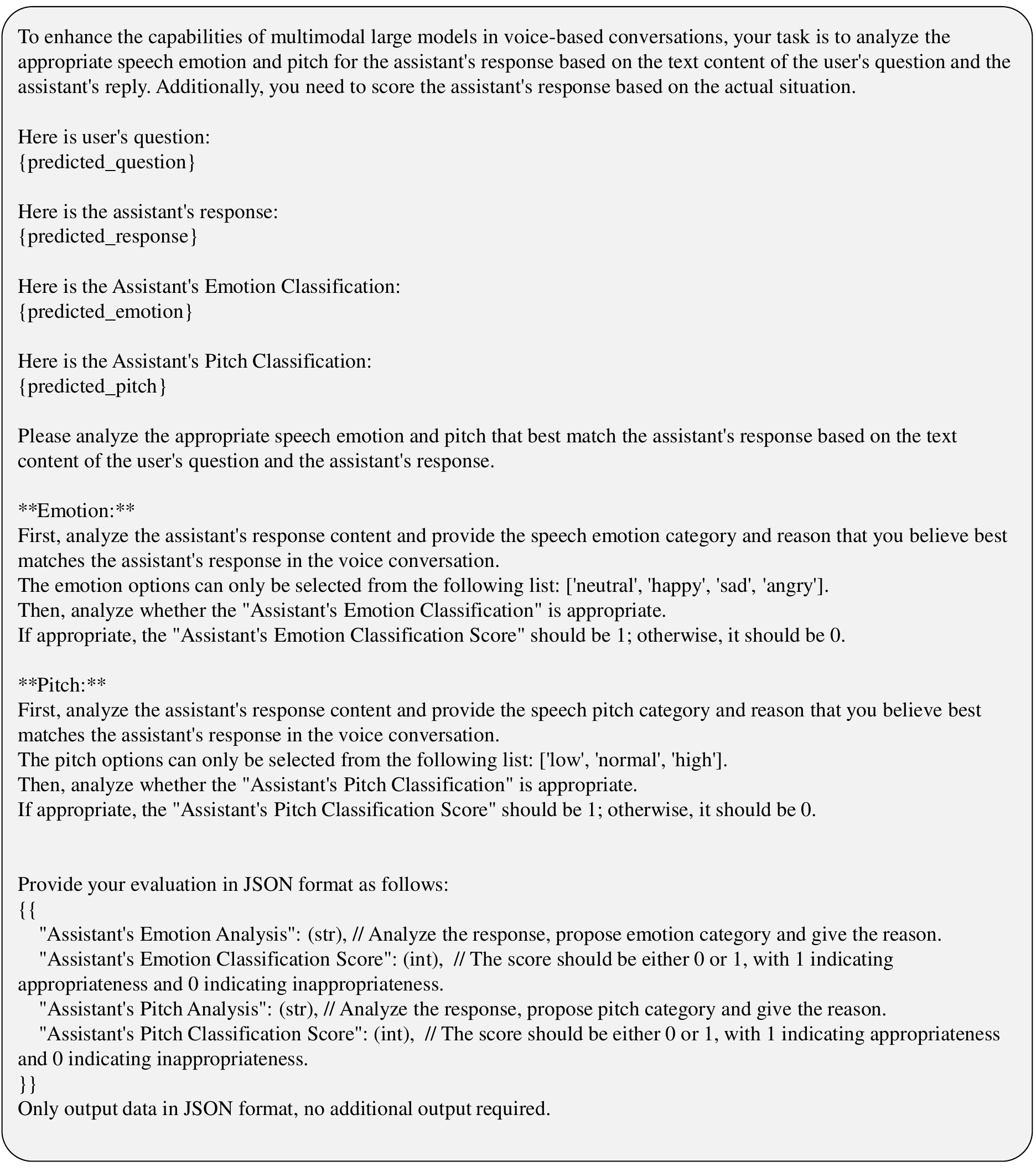}
\caption{\textbf{Prompt} used to obtain \textbf{Classification Accuracy of Style Label}.}
\label{fig:4o_style}
\end{figure}

\paragraph{Style label classification accuracy.} We use GPT-4o to decide whether the style predictions $\tilde{c_\text{emo}^{o}}, \tilde{c_\text{p}^{o}}$ are correct given the transcribed instruction $\tilde{x_T^{o^1}}$ and the predicted text response $\tilde{x_T^{o^2}}$. The prompt can be found in Fig. \ref{fig:4o_style}.

\paragraph{Emotion controllablity} of our U2S detokenizer is assessed by providing texts to LLM to generate corresponding units (\ie, TTS), which, along with the given conditional emotion labels, are then fed into our U2S detokenizer to synthesize speech.
We choose the female voice due to its large variation of styles. We evaluate on 4 commonly-seen emotions, \ie,``neutral'', ``happy'', ``sad'', and ``angry''. We synthesize 200 speech utterances for testing, with 50 utterances per emotion. 
The output speeches are analyzed by a Speech Emotion Recognition (SER) model named Emotion2Vec~\citep{ma2023emotion2vec}, which identifies the emotion with the greatest likelihood among these four emotions.



\begin{table}[t]
\begin{adjustbox}{width=0.55\linewidth, center}
\begin{tabular}{lcccc}
\toprule 
\multirow{2}{*}{Models} &\multicolumn{2}{c}{Librispeech (EN)} & \multicolumn{2}{c}{AISHELL-2 (ZH)} \\
& WER$\downarrow$ & TTS-WER$\downarrow$ & CER$\downarrow$ & TTS-CER$\downarrow$ \\
\midrule
Whisper-Large~\citep{radford2022robustspeechrecognitionlargescale} & 3.0 & - & - & - \\
Mini-Omni~\citep{xie2024mini} & 4.5 & - & - & - \\
\midrule
AnyGPT~\citep{zhan2024anygptunifiedmultimodalllm} & 8.5 & - & - & - \\
VITA~\citep{fu2024vitaopensourceinteractiveomni} & 8.1 & - & - & - \\
\midrule
\rowcolor{backcolor}
\textbf{\names-3B (ours)} & 5.4 & 5.8  & 16.6 & 17.5 \\
\rowcolor{backcolor}
\textbf{\names-7B (ours)} & 4.1 & 3.6  & 14.4 & 10.1 \\
\rowcolor{backcolor}
\textbf{\names-72B (ours)} & \textbf{2.9} & \textbf{3.5}  & \textbf{7.2} & \textbf{5.8} \\
\bottomrule
\end{tabular}%
\end{adjustbox}
\vspace{-3mm}
\caption{\textbf{Comparison on the ASR and TTS benchmarks.}}
\label{tbl_asr_tts}
\vspace{-2mm}
\end{table}


\subsection{Comparison with other Omni Models}
\label{app:omni compare}

Experimental results of ASR and TTS are reported in Table~\ref{tab:main_table} and~\ref{tbl_asr_tts}.
\textbf{\names} achieves significant improvements over other omni-modal models (\ie, AnyGPT and VITA), even surpassing its SLLM counterpart Mini-Omni~\citep{xie2024mini}, demonstrating the effectiveness of semantic-acoustic disentanglement and omni-modal mutual benefits.
For the first time, our \textbf{\names} obtains state-of-the-art performance on both the vision-language and speech benchmarks simultaneously.


\section{More on Vision-language Architecture}


\begin{table}[htbp]
  \centering
  \setlength{\tabcolsep}{1mm}
  \resizebox{\textwidth}{!}{%
    \begin{tabular}{ccc|cccccccccc}
    \toprule
    ViT   & ViT LR & Template & MME   & MMBench & SEED-Image & TextVQA & ChartQA & DocVQA & InfoVQA & OCRBench & ScienceQA-Img & AI2D \\
    \midrule
    Full  & 2$\times10^{-6}$  & QA    & \textbf{1928} & \underline{68.8} & 72.5 & \textbf{64.3} & 29.9 & \underline{45.2} & \underline{28.7}  & \textbf{495}   & \underline{76.3}  & 61.8 \\
    \rowcolor{backcolor} Half & 2$\times10^{-6}$ & QA     & 1838 & \textbf{71.3} & \underline{72.8}  & \underline{63.3} & \textbf{31.4}  & \textbf{46.0} & 28.5 & \underline{489}   & 76.1  & \textbf{63.7} \\
    Frozen  & 2$\times10^{-6}$  & QA     & 1887 & 68.8 & 72.2 & 61.3 & \underline{30.2} & 44.7 & 28.0 & 478   & 75.9 & 62.8 \\
    \midrule
    Half & 2$\times10^{-5}$  & QA     & 1833 & 68.3 & \textbf{73.1} & 62.2 & 27.8  & 42.4 & 27.3 & 461   & 74.7 & 63.4 \\
    Half & 2$\times10^{-6}$  & Plain  & \underline{1909} & 70.1 & 72.0 & 61.5 & 24.5 & 38.9 & \textbf{30.1} & 410   & \textbf{77.0} & 63.6 \\
    \bottomrule
    \end{tabular}%
    }
    \vspace{-3mm}
    \caption{\textbf{Ablation on the ViT configurations and templates for vision-language alignment.}}
    \label{tab:app_vl_exp}%
    \vspace{-3mm}
\end{table}%


This section analyzes the pre-training configurations of the vision encoder and the prompt template during Stage 2, as shown in Table.~\ref{tab:app_vl_exp}. Our final selection is the colored setting. We find that training the ViT model with half of the deeper parameters~\citep{chen2023sharegpt4v} with a learning rate of 2$\times10^{-6}$~\citep{liu2024llavanext} yields the best performance. Furthermore, we compare the QA template with the plain template~\citep{liu2023llava} and find that the QA template is superior for pre-training. 


\section{Limitations}

\paragraph{Duplex modeling.}

In the current version, \textbf{\names} can only process either visual/speech/text inputs or produce speech/text outputs at the same time. For a communication experience that mirrors human interaction, handling inputs and outputs simultaneously is crucial. 
Recent works like VITA \citep{fu2024vitaopensourceinteractiveomni} and LSLM~\citep{ma2024language} have begun to explore duplex modeling. 
VITA focuses on recognizing speech in noisy environments during the generation process to facilitate timely responses. LSLM, on the other hand, attempts to halt speech production when it detects a command or voice. Recently, a ground-breaking work named Moshi~\citep{kyutai2024moshi} develops a model supporting fully duplex modeling. The adeptness at simultaneously managing the information streams from both the user and the assistant allows Moshi to converse with human beings in real-time scenarios.

However, incorporating emotions into this duplex modeling presents additional challenges. Emotional states can fluctuate throughout a conversation, and how to generate appropriate responses given the user's previous and current emotional cues has not been thoroughly investigated. We will dive into this topic in the future work.

\vspace{-2mm}
\paragraph{Direct unit-to-unit generation.}
Although the speech units have served as the speech representation, they are predominantly adopted in conjunction with text-based assistance~\citep{zhang2023speechgpt}. 
However, the direct generation from unit to unit without text assistance is an area that has not been extensively explored. 
In \cite{lee2021textless}, speeches from the source language are directly translated into speech units of the target language for speech-to-speech translation. Similarly, \cite{nguyen2023generative} builds a language model directly on speech units, enabling spoken dialogue generation from raw audio. Both works develop models in speech-only data.

In the current version of \textbf{\names}, the text modality is integrated into the speech generation process to transfer textual knowledge to the speech modality, thereby enhancing the correctness of speech responses. In the future, we will strengthen the model's direct unit-to-unit generation capabilities to boost the speed of speech generation and augment the model's comprehension of speech modality.

\vspace{-2mm}
\paragraph{Vision configurations.}
Currently, we only utilize a single vision encoder pre-trained via a vision-language manner, while recent works have shown effectiveness by combining vision encoders pre-trained by different manners (\eg, self-supervised pre-training~\citep{liu2022task}) and architectures (\eg, MoE \citep{zhili2023task,liu2024mixture}).
We prioritize visual understanding in this work, while the incorporation of (controllable) visual generation~\citep{chen2023integrating,gao2023magicdrive,li2023trackdiffusion,wang2024detdiffusion,liu2023geomerasing,gao2024magicdrive3d,gao2024magicdrivedit} is also appealing to better empower \textbf{\names} in real-life applications~\citep{li2024automated,li2022coda,han2021soda10m,wu2024unified}.
Digging into its robustness towards noisy vision inputs~\cite{gou2025corrupted} is also an appealing direction.

\vspace{-2mm}
\paragraph{Comparison with Emu3~\cite{wang2024emu3}.}
Both our \textbf{\names} and Emu3 build end-to-end MLLMs with \textit{discrete tokenization}.
Emu3 discretizes \textit{visual data}, enabling visual understanding and generation, while our \textbf{\names} discretizes \textit{speech data} with a \textit{continuous visual encoder}, building an Omni-modal LLM with visual, text, and speech abilities simultaneously.


\section{Qualitative Results}
\label{App:Qualitative Results}

\begin{figure}[t]
\centering
\includegraphics[width=0.9\columnwidth]{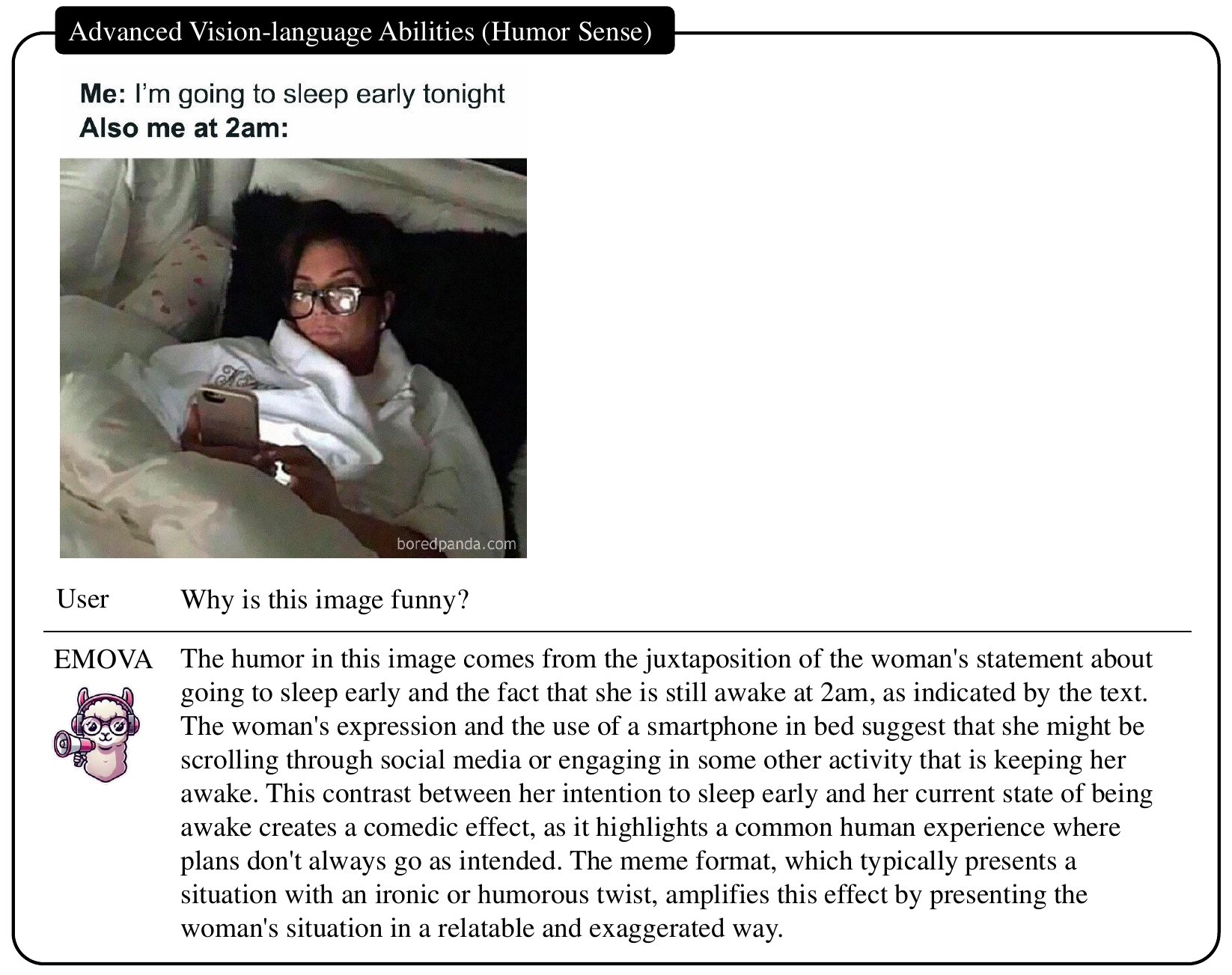}
\vspace{-3mm}
\caption{\textbf{\names} exhibits \textbf{advanced vision-language abilities in humor sense}.}
\vspace{-3mm}
\label{fig:case_vl_humor}
\end{figure}

\begin{figure}[t]
\centering
\includegraphics[width=\columnwidth]{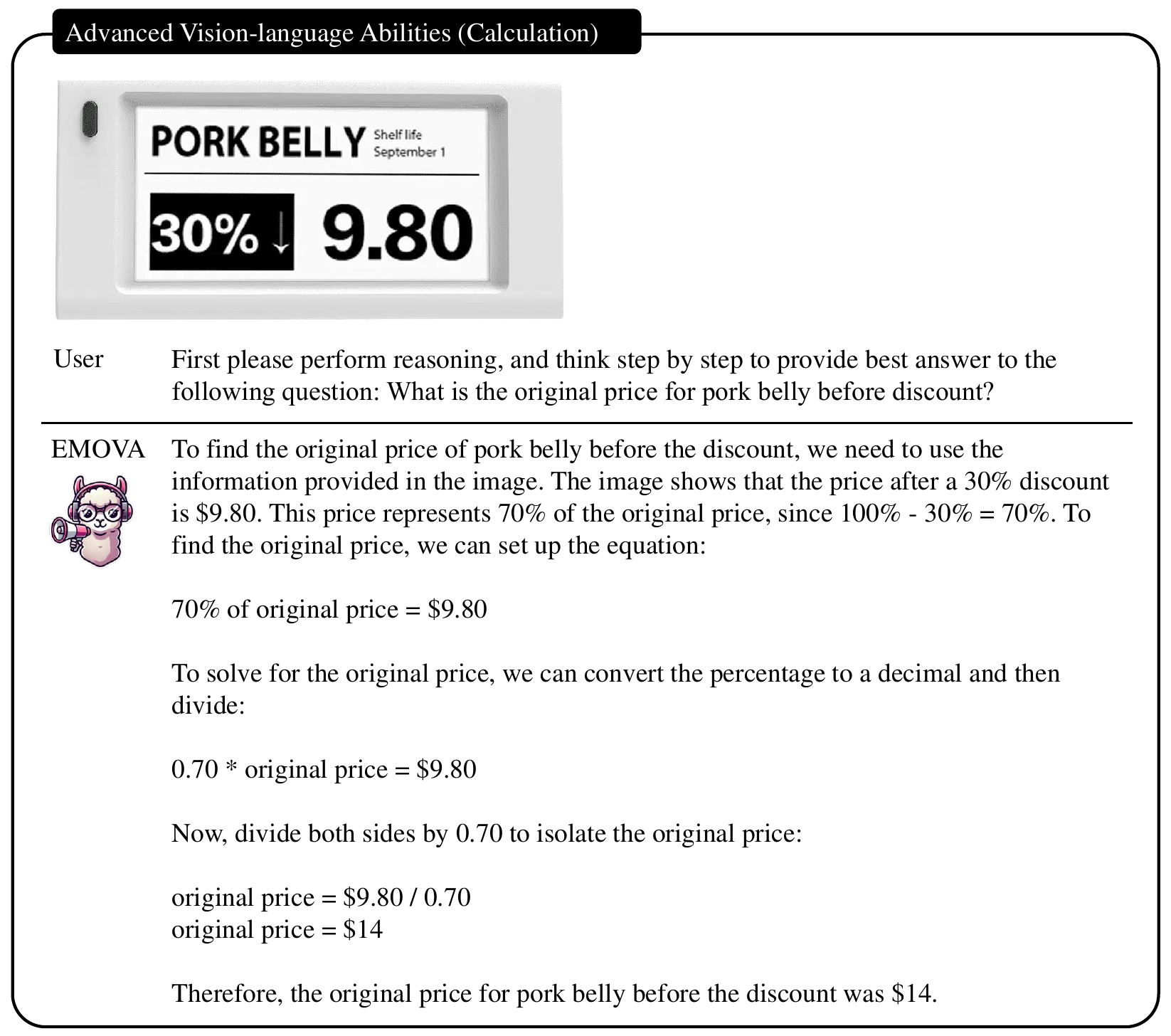}
\caption{\textbf{\names} exhibits \textbf{advanced vision-language abilities in numerical calculations}.}
\label{fig:case_vl_cal}
\end{figure}

\begin{figure}[t]
\centering
\includegraphics[width=0.7\columnwidth]{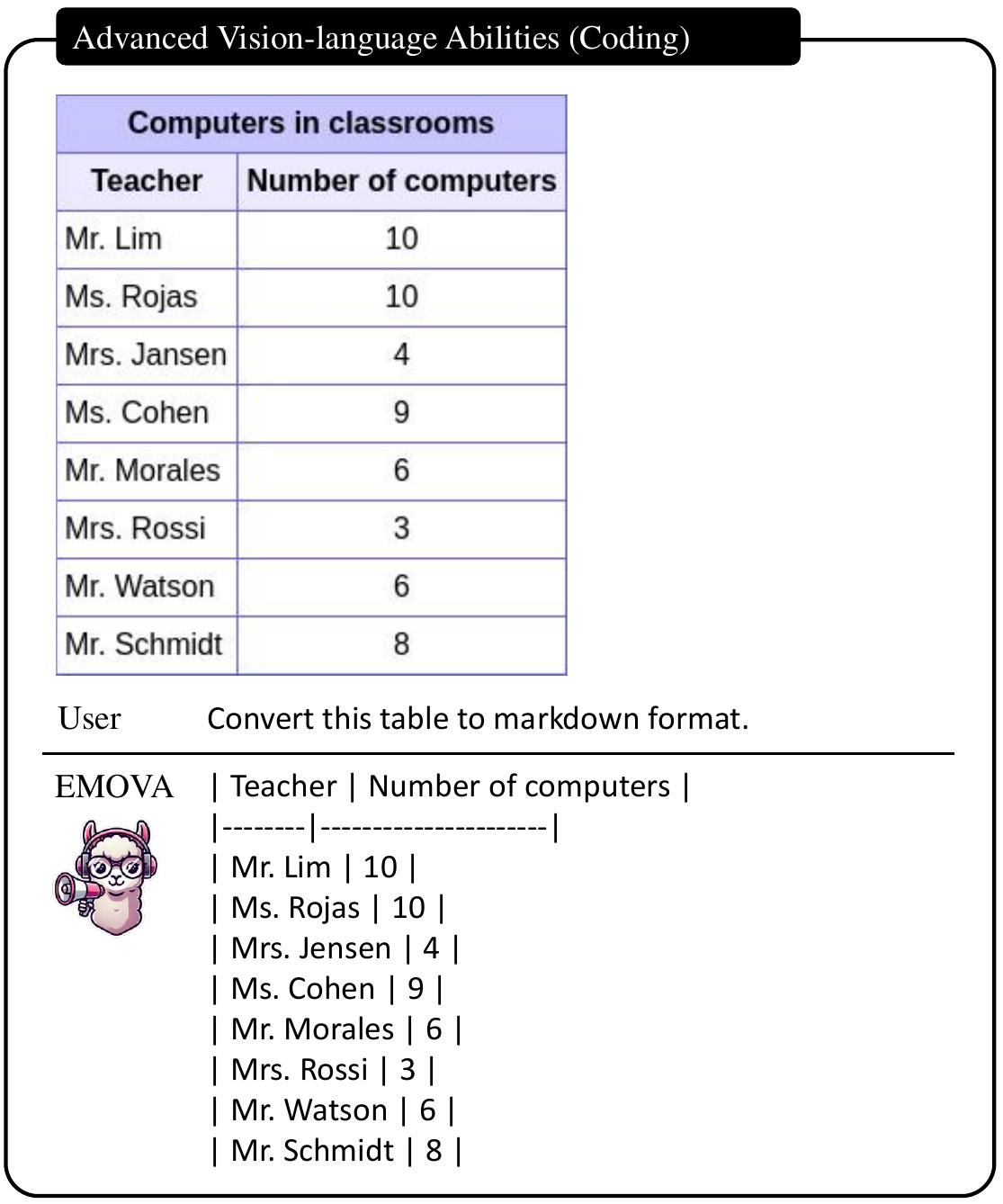}
\caption{\textbf{\names} exhibits \textbf{advanced vision-language abilities in coding}.}
\label{fig:case_vl_code}
\end{figure}

\begin{figure}[t]
\centering
\includegraphics[width=\columnwidth]{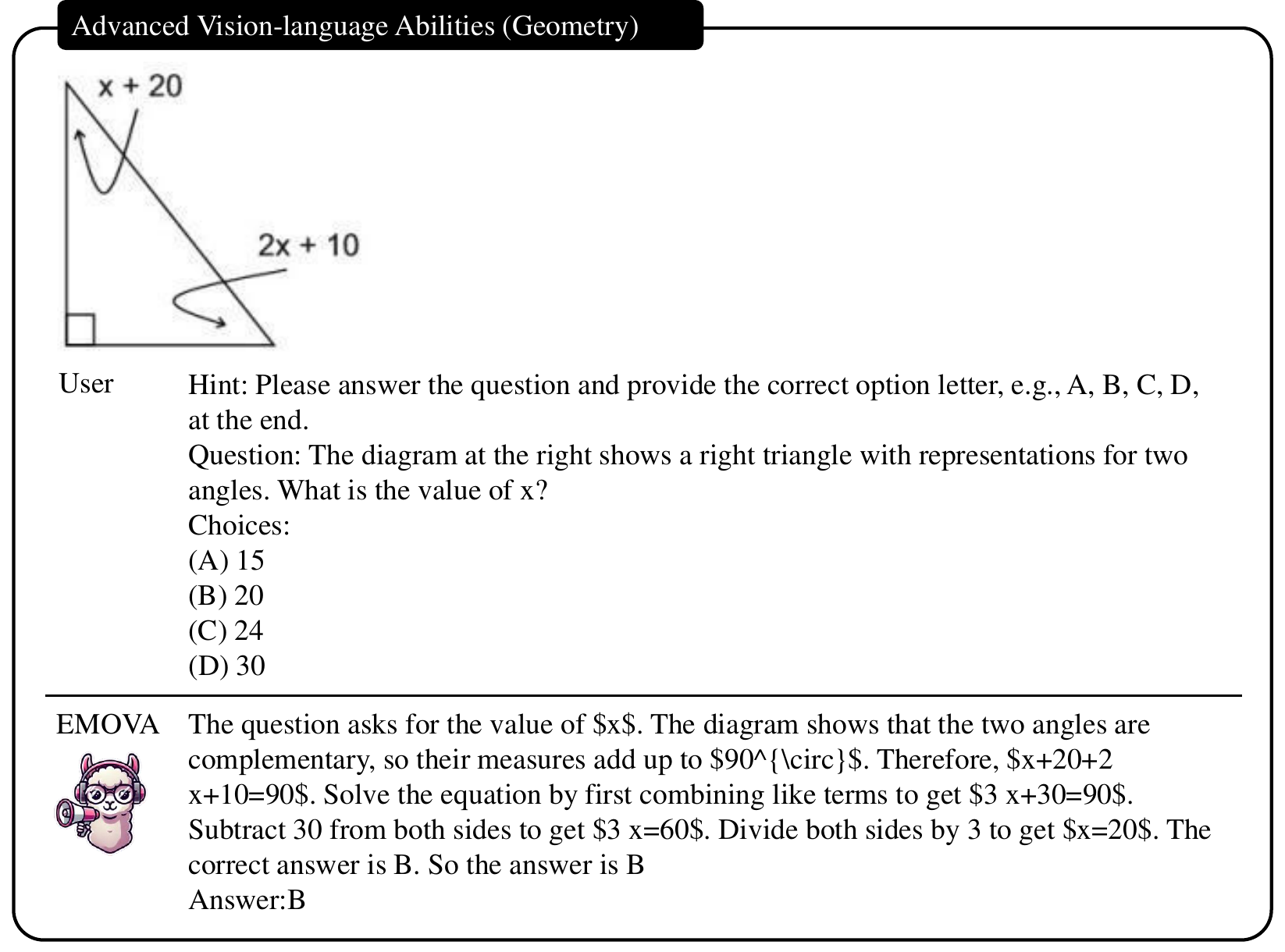}
\caption{\textbf{\names} exhibits \textbf{advanced vision-language abilities in math}.}
\label{fig:case_vl_math}
\end{figure}

\paragraph{Advanced Vision-language Abilities.}
Our \textbf{\names} exhibits advanced vision-language abilities in humor understanding (Fig. \ref{fig:case_vl_humor}), numerical calculations (Fig. \ref{fig:case_vl_cal}), coding (Fig. \ref{fig:case_vl_code}), geometry problem solving (Fig. \ref{fig:case_vl_math}). 

\begin{figure}[t]
\centering
\includegraphics[width=0.8\columnwidth]{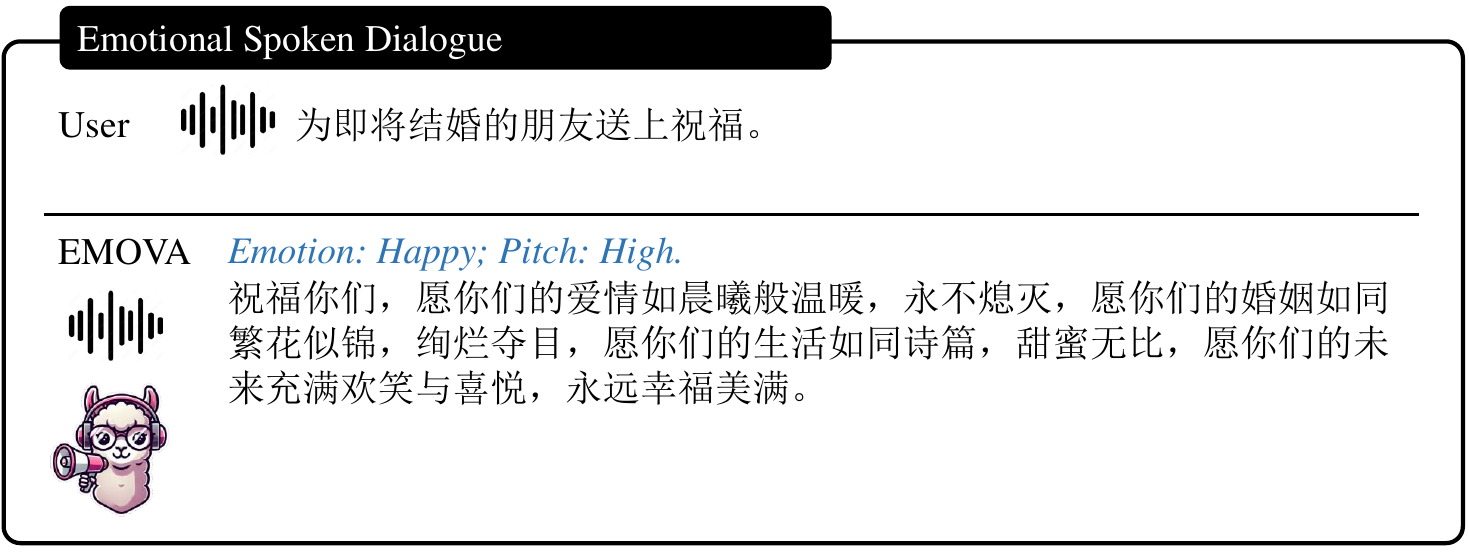}
\caption{\textbf{\names} engages in \textbf{spoken dialogue with a cheerful tone}.}
\vspace{-3mm}
\label{fig:case_emo_wedding}
\end{figure}

\begin{figure}[t]
\centering
\includegraphics[width=\columnwidth]{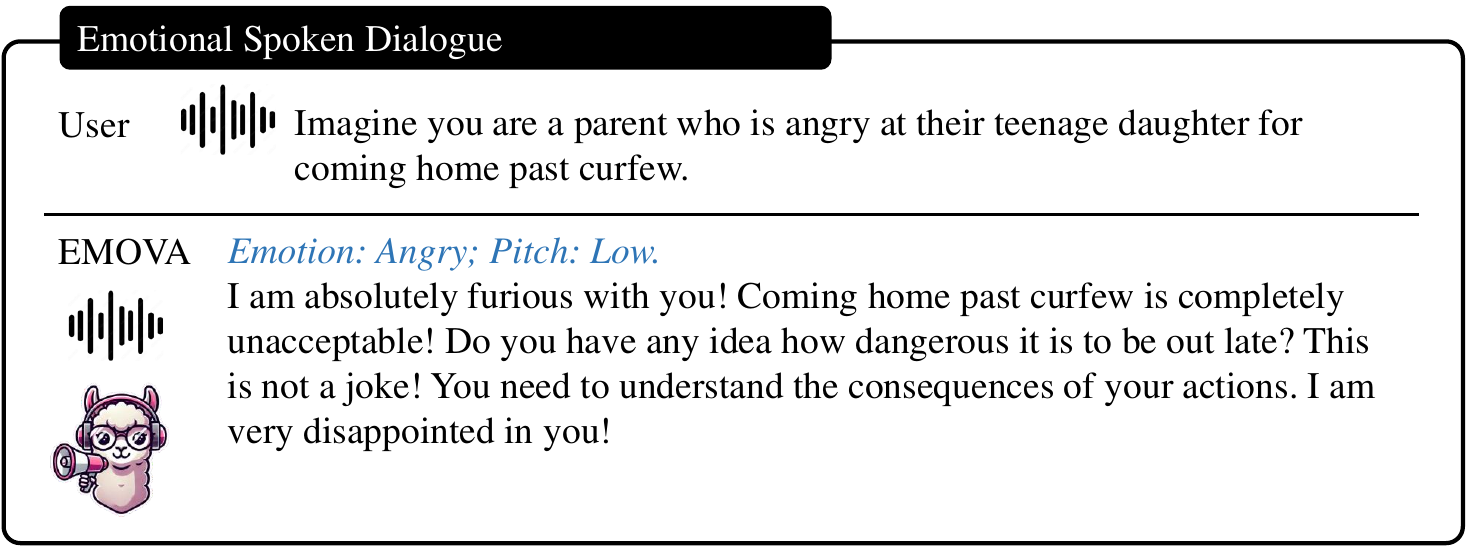}
\caption{\textbf{\names} engages in \textbf{emotional spoken dialogue}.}
\label{fig:case_emo_parent}
\end{figure}

\begin{figure}[t]
\centering
\includegraphics[width=\columnwidth]{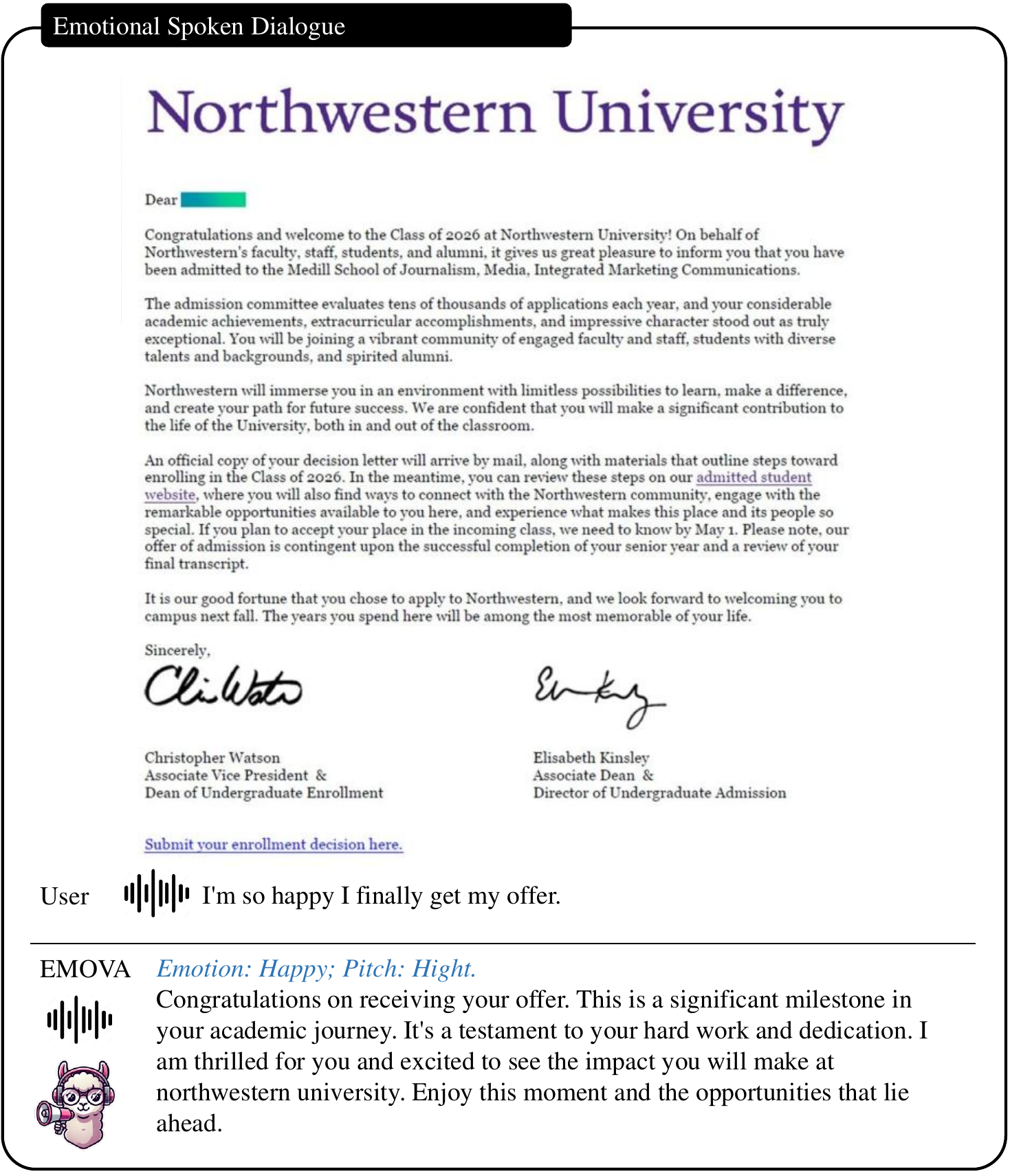}
\caption{\textbf{\names} engages in \textbf{omni-modal emotional spoken dialogue}.}
\label{fig:case_emo_offer}
\end{figure}

\paragraph{(Omni-modal) Emotional Spoken Dialogue.}
\textbf{\names} can engage in omni-modal emotional spoken dialogue (Figs. \ref{fig:case_emo_poem}, \ref{fig:case_emo_wedding}, \ref{fig:case_emo_parent}, \ref{fig:case_emo_offer}). For demonstration purposes, we present only the speech transcriptions\footnote{For speech files, please refer to our project page.} of the instruction and \names's response. The text highlighted in blue indicates the style labels predicted by \names, reflecting the emotion and pitch of generated speeches.

\begin{figure}[t]
    \vspace{-10mm}
    \centering
    \includegraphics[width=0.75\columnwidth]{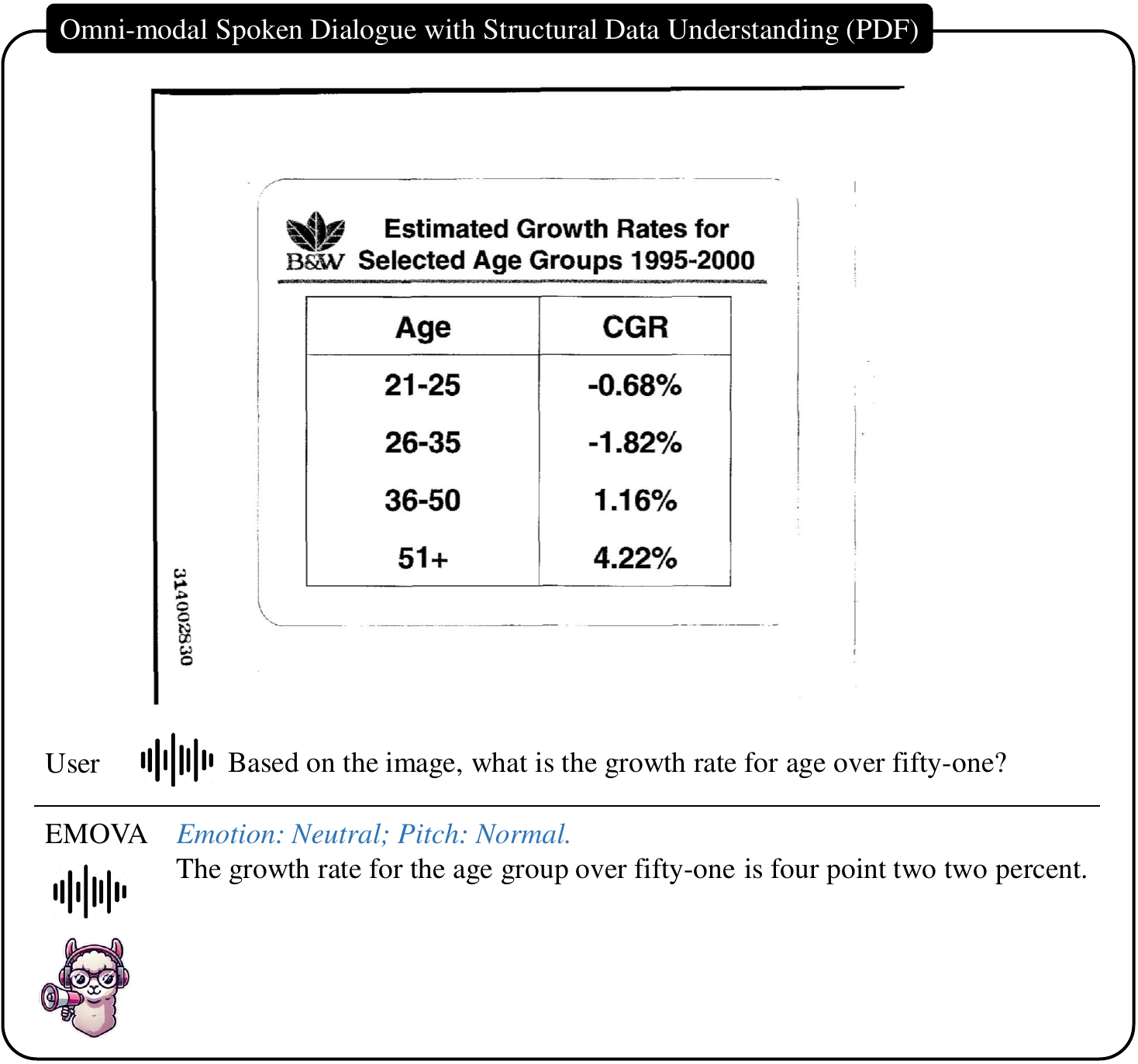}
    \caption{\textbf{\names} engages in omni-modal spoken dialogue with \textbf{structural data understanding}.}
    \label{fig:case_struc_speech_pdf}
    \vspace{-3mm}
\end{figure}

\begin{figure}[t]
\centering
\includegraphics[width=\columnwidth]{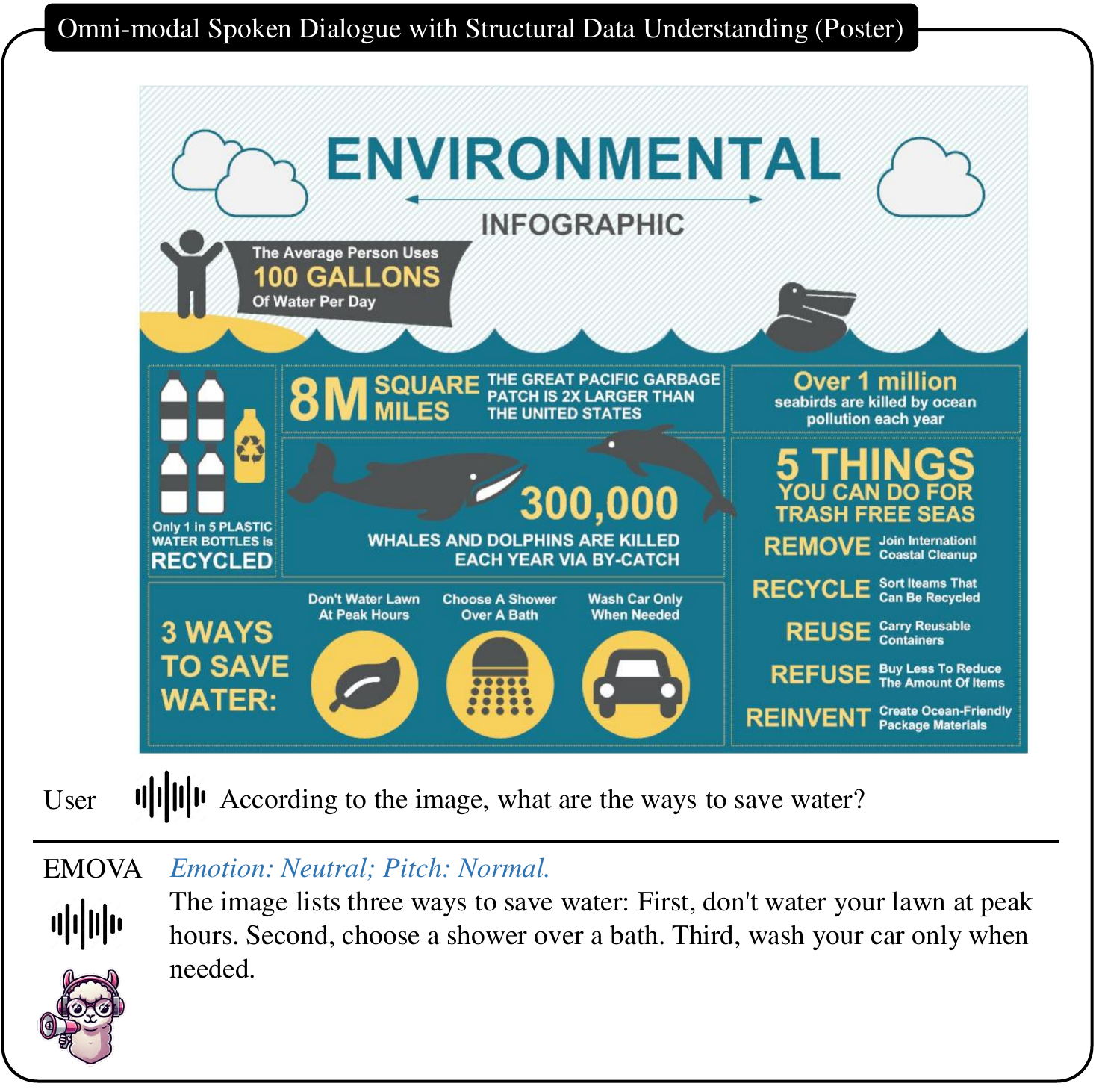}
\caption{\textbf{\names} engages in \textbf{omni-modal spoken dialogue with structural data understanding (\ie, poster)}.}
\label{fig:case_struc_speech_poster}
\end{figure}

\begin{figure}[t]
\centering
\includegraphics[width=\columnwidth]{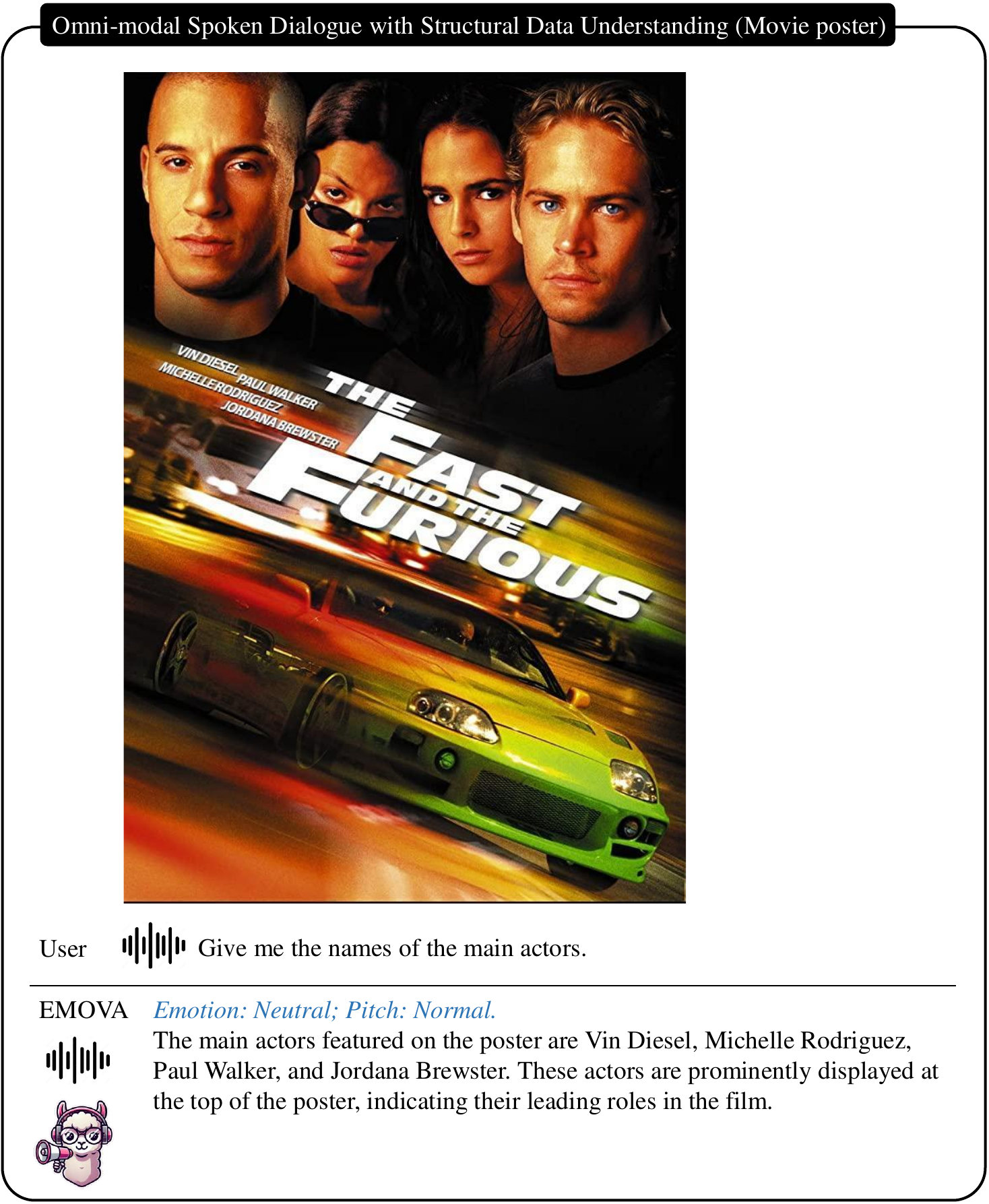}
\caption{\textbf{\names} engages in \textbf{omni-modal spoken dialogue with structural data understanding (\ie, movie posters)}.}
\label{fig:case_struc_speech_movie}
\end{figure}

\begin{figure}[t]
\centering
\includegraphics[width=0.75\columnwidth]{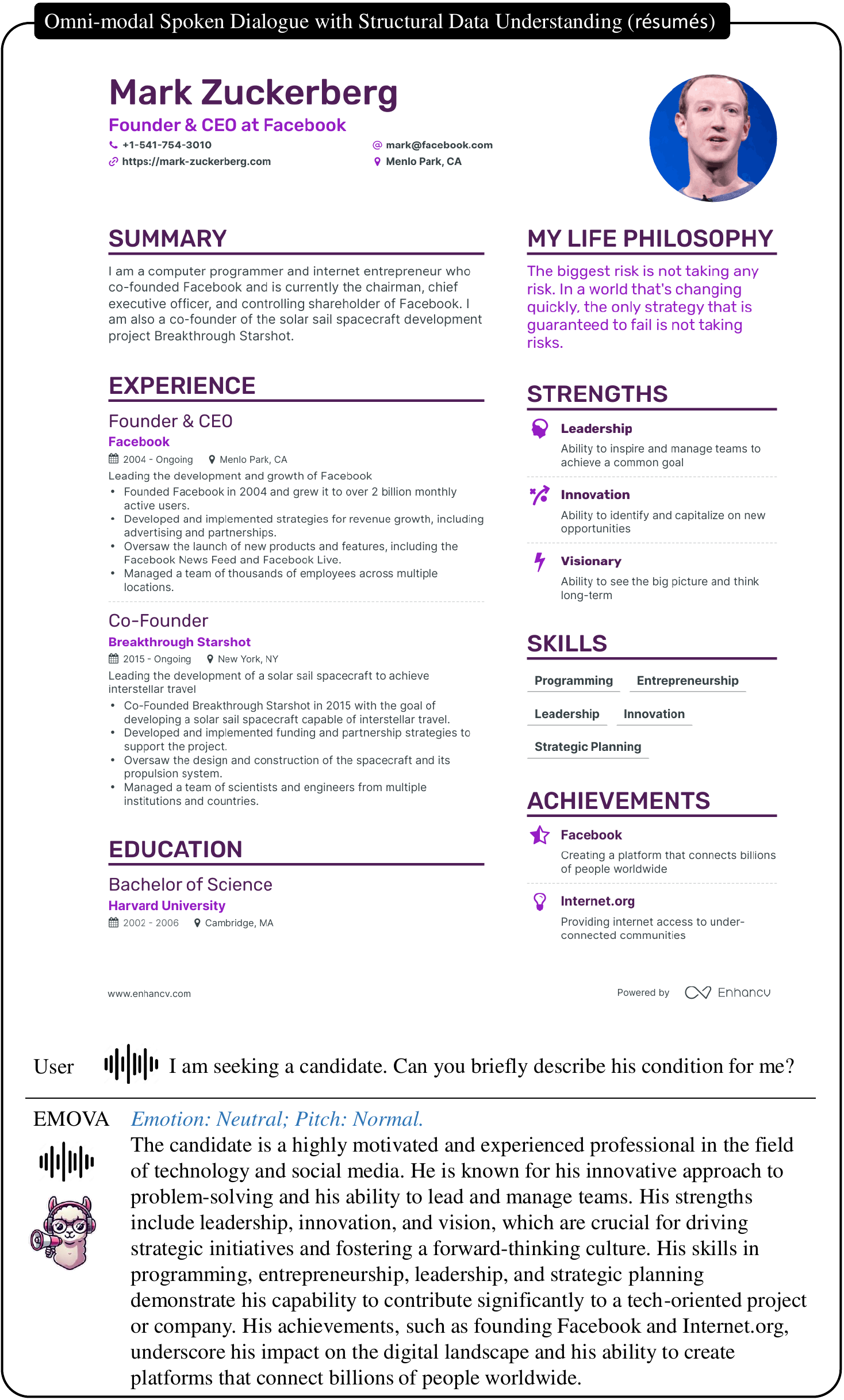}
\caption{\textbf{\names} engages in \textbf{omni-modal spoken dialogue with structural data understanding (\ie, résumé)}.}
\label{fig:case_struc_speech_cv}
\end{figure}

\begin{figure}[t]
\centering
\includegraphics[width=0.95\columnwidth]{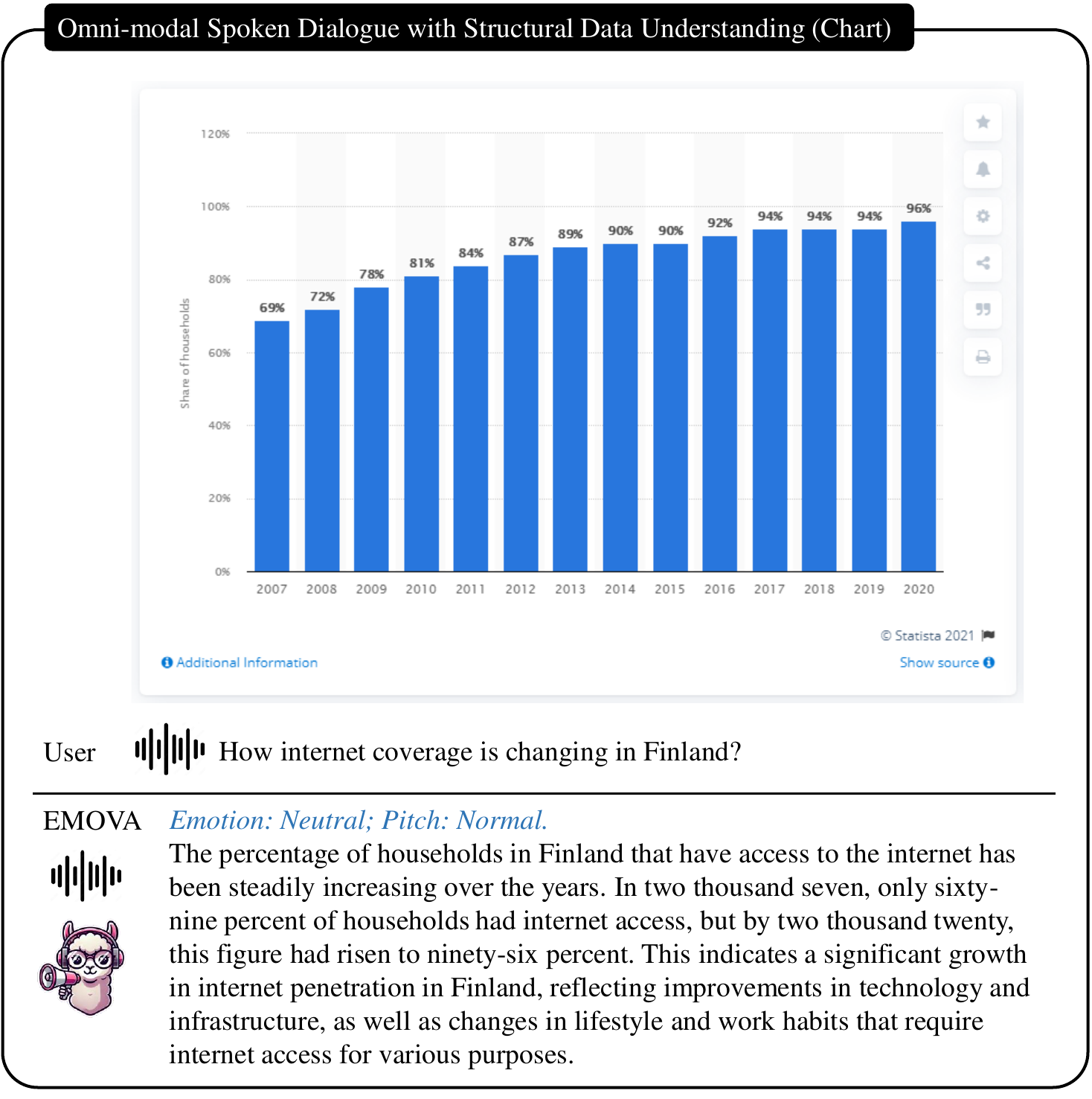}
\caption{\names engages in \textbf{omni-modal spoken dialogue with structural data understanding (\ie, chart)}.}
\label{fig:case_struc_speech_chart}
\end{figure}

\begin{figure}[t]
\centering
\includegraphics[width=0.95\columnwidth]{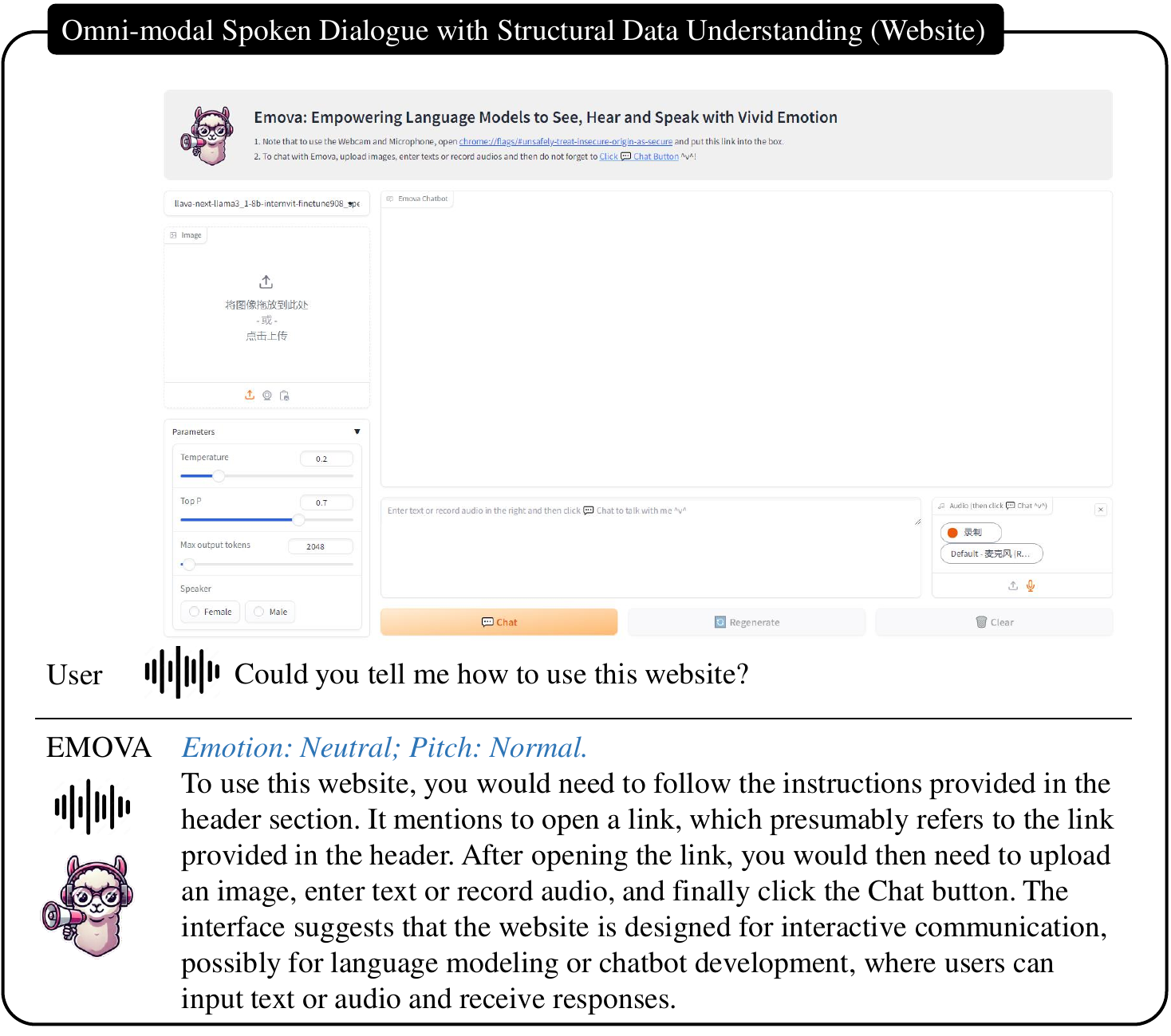}
\caption{\textbf{\names} engages in \textbf{omni-modal spoken dialogue with structural data understanding (\ie, website)}.}
\label{fig:case_struc_speech_website}
\end{figure}

\paragraph{Omni-modal Spoken Dialogue with Structural Data Understanding.}
\textbf{\names} can perform structural data understanding even in spoken dialogue. The data types include PDF files (Fig. \ref{fig:case_struc_speech_pdf}), movie posters (Figs. \ref{fig:case_struc_speech_poster} and \ref{fig:case_struc_speech_movie}), personal résumés (Fig. \ref{fig:case_struc_speech_cv}), charts (Fig. \ref{fig:case_struc_speech_chart}), and websites (Fig. \ref{fig:case_struc_speech_website}). Remarkably, \names is not specifically trained on spoken dialogue data involving these data types. This suggests that our chain of modality data introduced in Sec. \ref{sec:ommi instruction tuning} effectively activates both vision-language abilities and speech understanding acquired during the text-centric alignment stage.

